\documentclass[10pt,twocolumn,letterpaper]{article}
\usepackage[pagenumbers]{cvpr}

\usepackage{amsmath}
\usepackage{amssymb}

\usepackage{booktabs}
\usepackage[toc,page]{appendix}

\usepackage{array}
\usepackage{multirow}
\usepackage{algorithmic}
\usepackage[linesnumbered,ruled,vlined]{algorithm2e}
\usepackage{amsfonts}
\usepackage{wrapfig}
\usepackage{graphicx}
\usepackage{pifont}
\usepackage{multirow}
\usepackage{setspace}

\newcommand*{\affaddr}[1]{#1}

\newcommand*{\email}[1]{\texttt{#1}}
\usepackage[pagebackref,breaklinks,colorlinks]{hyperref}

\usepackage{subfloat}
\graphicspath{{./Figures/}}
\DeclareGraphicsExtensions{.pdf,.jpg,.png}

\usepackage{cleveref}
\crefformat{equation}{Eqn. (#2#1#3)}
\crefformat{figure}{Fig. #2#1#3}
\crefformat{section}{Sec. #2#1#3}
\crefformat{table}{Tab. #2#1#3}

\def\ie{\textit{i.e.}\xspace}

\def\etc{\textit{etc.}\xspace}
\def\eg{\textit{e.g.}\xspace}

\begin{document}

\title{Relational Representation Learning in Visually-Rich Documents}

\author{
  Xin Li\textsuperscript{\thanks{Equal contribution.}} \quad Yan Zheng\textsuperscript{*}  \quad Yiqing Hu\textsuperscript{\thanks{Contact person.}} \quad Haoyu Cao \\ Yunfei Wu \quad Deqiang Jiang \quad Yinsong Liu \quad Bo Ren\\   
  \affaddr{Tencent YouTu Lab} \quad\\
  \email{\small \{fujikoli, neoyzheng, hooverhu, rechycao, marcowu, dqiangjiang, jasonysliu, timren\}@tencent.com}
}
\maketitle

\begin{abstract}
Relational understanding is critical for a number of visually-rich documents (VRDs) understanding tasks.
Through multi-modal pre-training, recent studies provide comprehensive contextual representations and exploit them as prior knowledge for downstream tasks.
In spite of their impressive results, we observe that the widespread relational hints (\eg, relation of key/value fields on receipts) built upon contextual knowledge are not excavated yet.
To mitigate this gap, we propose \textbf{DocReL}, a \textbf{Doc}ument \textbf{Re}lational Representation \textbf{L}earning framework.
The major challenge of DocReL roots in the variety of relations.
From the simplest pairwise relation to the complex global structure, it is infeasible to conduct supervised training due to the definition of relation varies and even conflicts in different tasks.
To deal with the unpredictable definition of relations, we propose a novel contrastive learning task named Relational Consistency Modeling (RCM), which harnesses the fact that existing relations should be consistent in differently augmented positive views.
RCM provides relational representations which are more compatible to the urgent need of downstream tasks, even without any knowledge about the exact definition of relation.
DocReL achieves better performance on a wide variety of VRD relational understanding tasks, including table structure recognition, key information extraction and reading order detection.
\end{abstract}

\section{Introduction}
\label{sec:introduction}

Visually-Rich Documents (VRDs) understanding aims to automatically analyze and extract key information from scanned / digital-born documents, such as insurance quotes, business emails, and financial receipts.
Certain relational understanding tasks already play the important role in building electronic archives and developing office automation, which include table structure recognition~\cite{qasim2019rethinking,raja2020table,liu2021show,long2021parsing,liu2021neural}, key information extraction~\cite{hong2021bros,zhang2020trie,hwang2020spatial} and reading order detection~\cite{wang2021layoutreader}, \etc
Owing to their huge potential, VRDs understanding has attracted increasing attention in the multimedia community.
Recent studies~\cite{xu2020layoutlm,xu2020layoutlmv2,hong2021bros,li2021structext,li2021structurallm} usually follow a two-step pre-training then fine-tuning regime.
With cutting-edge model designs, the pre-training step jointly models a multi-modal interaction between text, layout, and image, then produces comprehensive representations for fine-tuning specific downstream tasks.

\begin{figure}[t]
  \centering
  \includegraphics[width=0.48\textwidth]{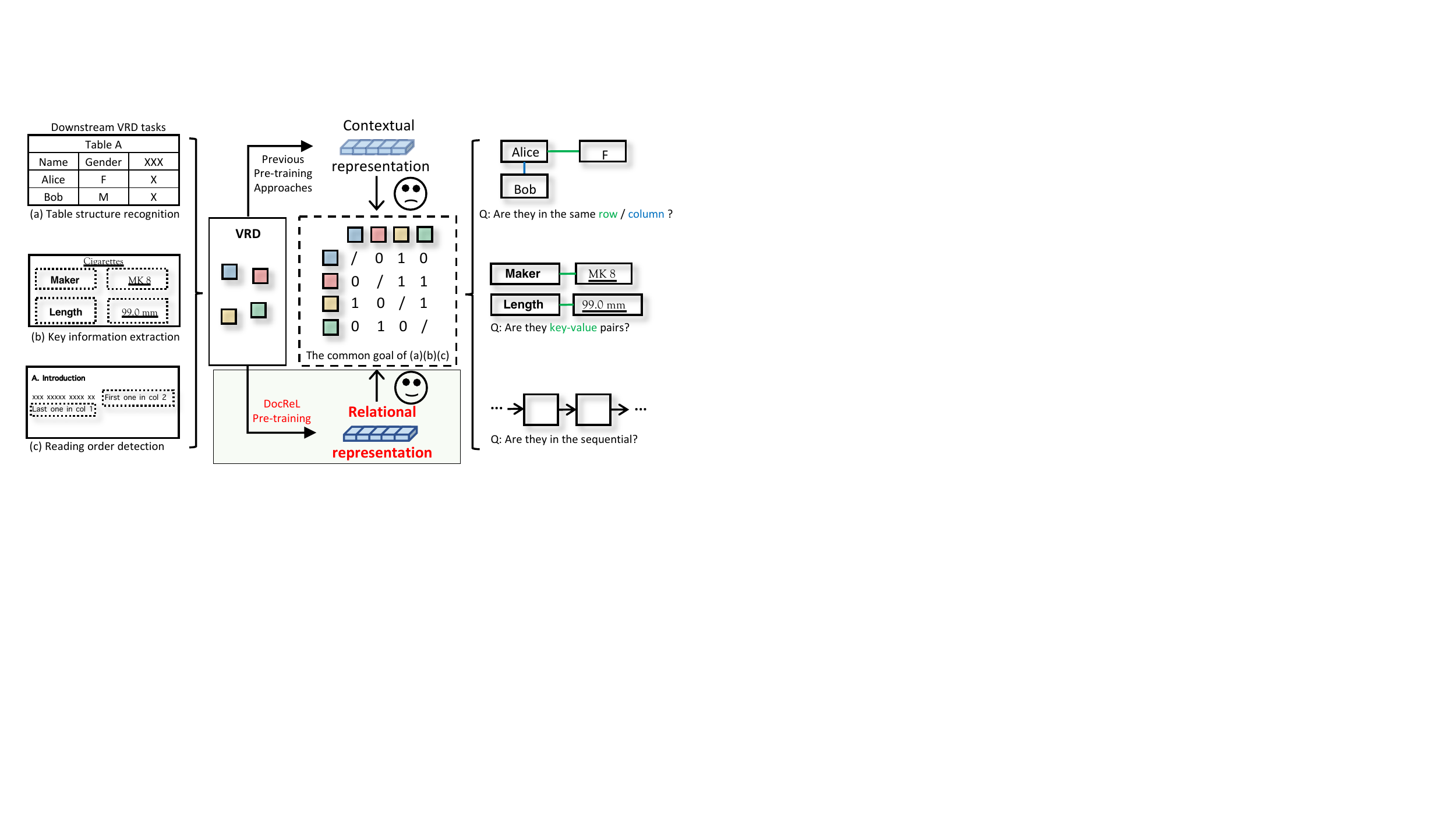}
    \caption{DocReL is a pre-training framework that facilitates typical VRDs understanding missions.
    Despite their goals varying, the common solution relies on analyzing the relation formed by semantic entities.
    DocReL firstly offers the relational feature representations that are more compatible to these tasks.
    }
  \label{fig:01}
\end{figure}

\begin{figure*}[t]
  \centering
  \includegraphics[width=1\textwidth]{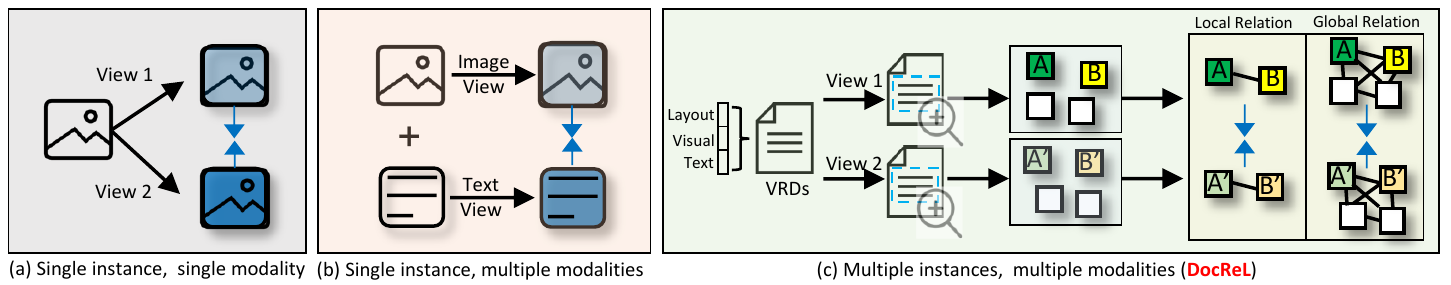}
  \caption{Comparison between DocReL and previous contrastive learning paradigms.
  DocReL exploits semantic entities with multi-modal features as the basic unit, and its goal is to learn relational representations in both local and global perspective.
  The blue arrows indicate feature comparison in contrastive learning and the colored blocks in (c) are positive views of the original entities.
  }
  \label{fig:02}
\end{figure*}

However, there remains a big gap between the representations provided by existing approaches and the demand of downstream tasks.
To provide comprehensive representations, recent schemes mostly exploit self-supervised pre-training tasks for they are expected to profit from common VRD collections without annotation.
Apart from the widely used cloze task~\cite{xu2020layoutlm} inspired by BERT~\cite{devlin2018bert},
mainstream self-supervised tasks usually learn representations by aligning multi-modal features of VRDs~\cite{xu2020layoutlmv2,li2021structurallm,appalaraju2021docformer,wu2021lampret,gu2021unidoc}.
These tasks perform well in understanding the multi-modal nature of VRDs and capturing the contextual knowledge.
However, they are inadequate to express a simple relation between pairwise semantic entities shown in~\cref{fig:01}, let alone the complex global structures of VRDs.
Under this circumstance, careful fine-tuning is necessary to satisfy the needs of downstream tasks.
Based on this observation, integrating the relational knowledge in advance, in other words, having the relational feature representations, will no doubt ease the burden of the fine-tuning stage.

To mitigate this gap, we propose DocReL, the first relational representations learning framework for VRDs understanding.
To realize DocReL, the major challenge comes from the variety of relations.
Even the definition of a straightforward pairwise relation may vary and even conflict in different tasks.
For instance, as shown in~\cref{fig:01}~(a), two entities ``Alice'' and ``Bob'' are in the same column.
Here entities are meaningful text sequences with corresponding layouts and images in VRDs.
In this case, their binary relation could be annotated as label ``1'' which indicates relevance.
On the other hand, they are not in the same row, thus their binary relation may be annotated as label ``0'' which indicates irrelevance.
To overcome this problem, we apply contrastive learning to study relational representations by maximizing agreement between differently augmented views of the same ``relation''.
As the consequence, the unpredictable relation is no longer an obstacle.

The design of our contrastive relational learning framework is not a trivial task.
Rather than directly regarding the raw input (\ie, a VRD image) as the basic instance~\cite{chen2020simple}, DocReL focuses on the finer-grained entities.
These entities form a wide range of relations, from the simplest pairwise relation to the complex global structure.
In addition, compared with previous approaches that leverage uni-modal visual data~\cite{chen2020simple} (\cref{fig:02}~(a)) or visual-text pairs~\cite{radford2021learning,yu2022commercemm} (\cref{fig:02}~(b)), VRDs offer multi-modal triplets that consist of visual, text, and layout, as shown in~\cref{fig:02}~(c).
Such an intricate combination enforces DocReL to 1) exactly extract the multi-modal feature of entities and 2) comprehensively model the local and the global relation formed by entities.

We address the above challenges with the following pipeline.
Firstly, we introduce a feature extraction system to learn cross-modal knowledge, which is properly organized to extract entity-level features.
In addition, we propose a novel pre-training task named Relational Consistency Modeling (RCM).
As relation modification is unforeseeable in negative augmentation views, 
RCM follows the prior of BYOL~\cite{grill2020bootstrap} to exploit only positive augmentation views.
RCM is formed by local relational consistency and global relational consistency, in which the former focuses on the relation between pairwise entities and the latter concentrates on the global structure of VRDs.
Finally, we customize a relational feature instantiation module to convert the relational representations to an easily accessible format for relation understanding tasks.

The contributions of this paper are summarized as follows:
\begin{enumerate}
\item DocReL firstly provides the relational feature representations, which is in stark contrast to previous works that provide contextual feature representations.
Ablation studies demonstrate that the relational feature representations are more compatible with downstream relation understanding tasks.
\item We coin a novel self-supervised pre-training task named RCM.
It is unrestricted to specific relations, taking into effect using only raw VRDs without annotation as the input.
RCM is also easily accessible to other pre-training frameworks for VRDs.
\item Experimental results show that DocReL achieves better performance on mainstream relational understanding tasks of VRDs, which include 1) SciTSR~\cite{chi2019complicated}, ICDAR-2013~\cite{gobel2013icdar} and ICDAR-2019~\cite{gao2019icdar} in table structure recognition, 2) FUNSD~\cite{jaume2019funsd} and CORD~\cite{park2019cord} in key information extraction and 3) ReadingBank~\cite{wang2021layoutreader} in reading order detection.
\end{enumerate}

\section{Related Works}
\label{sec:relatedwork}

DocReL is a pre-training framework that focuses on learning relational representations for mainstream relational understanding tasks on VRDs.
The core design of DocReL is a pre-training task named Relational Consistency Modeling (RCM), which is motivated by the recent progress of contrastive learning.
In this section, we briefly introduce these related approaches.

\subsection{Relational Understanding in VRDs}
Relational understanding has been intensively studied in a long history.
Pioneer works concentrate on identifying the relation between two named entities within the same sentence.
Based on them, sentence-level and document-level relational understanding~\cite{zeng2015distant,feng2018reinforcement,zhang2021document} appeals to researchers as it enables long-range dependency extraction throughout an entire document.
Recently, the flourishing of multi-modal pre-training provides a new perspective.
Rather than purely text-level relations in conventional works, relations in VRDs derive from a multi-modal combination of text, layout and image, wherein each plays its own role in building relations.
Inspired by natural language processing systems~\cite{devlin2018bert,dai2019transformer}, the solution of VRDs understanding usually follows a two-step pre-training then fine-tuning regime.

For the pre-training step, a feature extraction system is firstly established to jointly model interactions among text, layout and image.
The following pre-training tasks are utilized to enhance the multi-modal feature representation.
For the fine-tuning step, the representation generated beforehand is utilized for tuning downstream tasks.
Typical downstream tasks in VRDs are key information extraction~\cite{hong2021bros,zhang2020trie,liu2019graph}, reading order detection~\cite{wang2021layoutreader} and table structure recognition~\cite{long2021parsing}, \etc
Despite their objectives varying, they could be considered as relational understanding tasks.

\subsection{Pre-training Tasks in VRDs} 
In order to learn a better multi-modal feature representation, various tasks have been introduced in the pre-training of VRDs.
Notable tasks are designed to align multi-modal features~\cite{xu2020layoutlmv2,li2021structurallm,appalaraju2021docformer,wu2021lampret,gu2021unidoc} or study the local context~\cite{xu2020layoutlm,devlin2018bert}.
Most of them are self-supervised, and are expected to profit from common text corpus and image collections.
Currently, contrastive learning becomes a new hot spot in designing self-supervised tasks.
SimCLR~\cite{chen2020simple,chen2020big} learns representations by maximizing agreement between differently augmented views of the same image.
To avoid the dependence on large batch size, BYOL~\cite{grill2020bootstrap} trains an online network to predict the target network representations of the same image under a different augmented positive view. 
This target network is updated with a slow-moving average of the online network. 
The MoCo series~\cite{he2020momentum,chen2020improved,chen2021empirical} build a dynamic dictionary with a queue and a moving-averaged encoder, which enables building a large and consistent dictionary on-the-fly that facilitates contrastive unsupervised learning.
In these schemes, the basic instance is directly the inputted image.
To better understand multi-media information like advertisement or commerce topics, recent works~\cite{yu2022commercemm,li2021align,radford2021learning} augment image-text pairs or even a flexible combination of text, image and other valuable data like query and click.
Compared with approaches that exploit the raw input (\eg, image, image-text pairs), the basic instance in DocReL is a semantic entity.
Abundant entities in VRDs form complex connections.
DocReL hunts cross-modality clues among them to understand their connections, from the simplest pairwise relation to the complex global structure.
Our proposed DocReL focuses on learning better relational representations, which are expected to benefit all relational understanding tasks mentioned above.

\section{Approach}
\label{sec:approach}

\begin{figure*}[t]
  \centering
  \includegraphics[width=.95\textwidth]{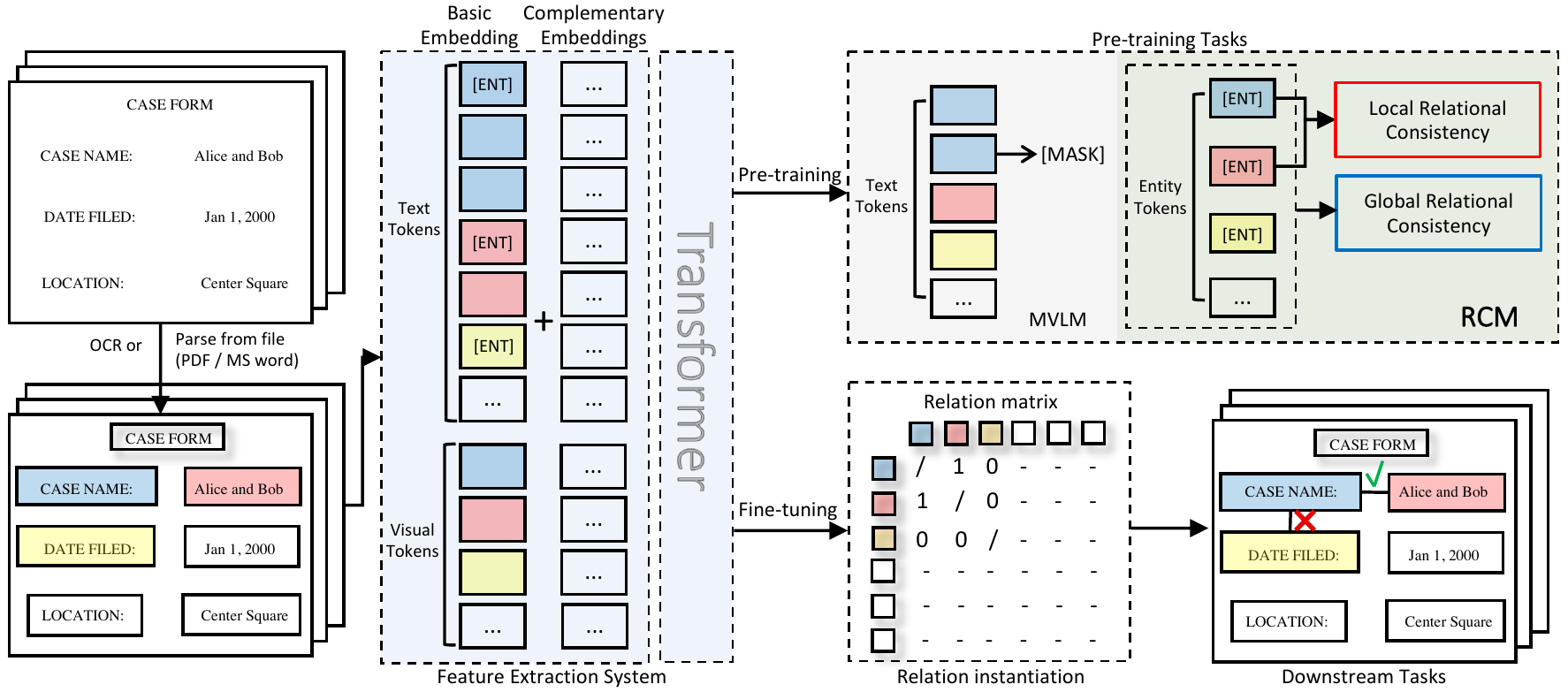}
  \caption{System overview of DocReL.
  DocReL consists of three major components, namely 1) a feature extraction module that is properly organized to extract representation of entities, 
  2) two pre-training tasks that guide the system to learn relational feature representations, where our proposed Relational Consistency Modeling (RCM) plays the key role, and
  3) a relational feature instantiation module that generates a unified relation matrix to facilitate downstream relational understanding tasks.
  }
  \label{fig:03}
\end{figure*}

\cref{fig:03} shows the overall architecture of DocReL. 
The input is an image with known entity fields.
As an entity is a meaningful text sequence with a corresponding layout and image, it could be a sentence to be sorted in reading order detection tasks~\cite{wang2021layoutreader}, could be a table cell to be located in table structure recognition tasks~\cite{liu2021show,liu2021neural,long2021parsing}, and also could be a key / value field to be identified in key information extraction tasks~\cite{hong2021bros,zhang2020trie}.
These entity fields are extracted either from preconditioned OCR results or parsing results from electronic formats like PDF or Microsoft Word.
To keep a fixed input format, these entities are sorted beforehand according to the top-left to bottom-right order~\cite{li2021structext}.

The DocReL system consists of three modules.
Firstly, the feature extraction module extracts a multi-modal feature from entities, then this feature embedding is fed into the Transformer~\cite{vaswani2017attention} network for feature fusion (\cref{sec:feature-embedding}).
In addition, two pre-training tasks are leveraged to learn relational representations: 1) Relational Consistency Modeling (RCM) which plays the key role for capturing the relational feature representation (\cref{sec:pretrain-task}), and 2) Masked Visual Language Modeling (MVLM) which is popular in VRDs understanding designs to learn better in the language side with the cross-modality clues~\cite{xu2020layoutlm}.
Finally, the relation instantiation module generates a relation matrix via the relational representations, which are the general input for downstream relational understanding tasks (\cref{sec:downstream}).

\subsection{Feature Extraction System}
\label{sec:feature-embedding}

Inspired by recent VRDs pre-training frameworks~\cite{xu2020layoutlm,wu2021lampret,xu2020layoutlmv2,li2021structext}, 
DocReL exploits multiple feature embeddings, which consists of 1) text-token embedding, 2) visual embedding, 3) layout embedding, 4) segment embedding and 5) modality embedding, as shown in~\cref{fig:03}.
Among them, the first two embeddings are fundamental and the others are complementary.
We carefully organize the embedding system to facilitate the feature extraction of each entity, which is the basic unit in learning relational representations.

To get the text-token embedding, WordPiece~\cite{wu2016google} is used for tokenization. 
For visual embedding, we firstly crop the corresponding image part according to its bounding box, then send it to ResNet-50~\cite{he2016deep} with FPN~\cite{lin2017feature} for visual feature extraction.
Notice that the visual embedding is entity-level, thus multiple text-token embeddings correspond to a same visual embedding token.
These two basic embeddings are concatenated in the vertical dimension, as shown in~\cref{fig:03}.
We add an [ENT] token to separate text tokens of each entity.
By virtue of this setting, entity-level features could be easily extracted by directly selecting the [ENT] token, which is already the fused multi-modal feature of entities through the Transformer.
A more detailed description of entity-level feature extraction in complex scenarios is given in supplementary material.

\begin{figure*}[t]
  \centering
  \includegraphics[width=1\textwidth]{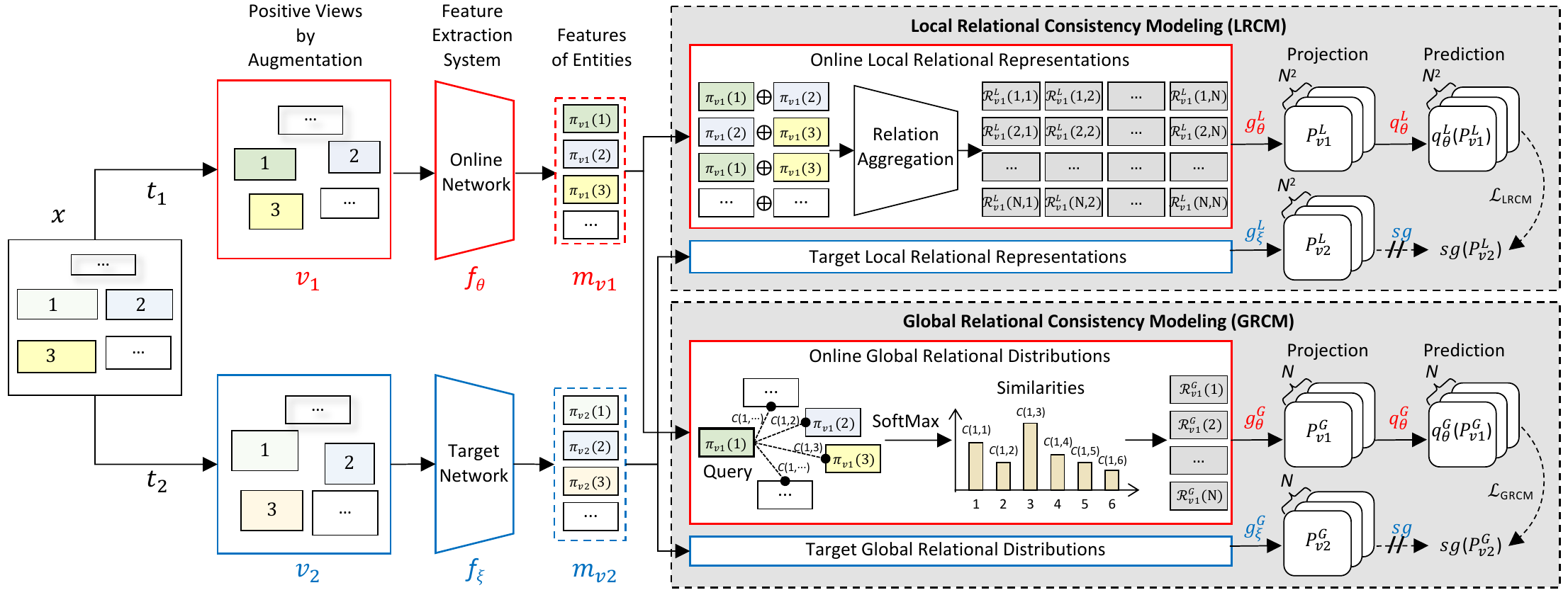}
  \caption{An illustrative view of Relational Consistency Modeling (RCM). 
  Two differently augmented positive views are created and fed into the online and the target feature extraction systems separately. 
  LRCM guarantees the consistency of local relational representations between pairwise entities, while GRCM keeps the consistency of the global relational distribution built by all entities. 
  During pre-training, parameters of the target networks $\xi$ are updated by an exponential moving average of the online networks' parameters $\theta$, and $sg$ represents the stop-gradient.
  Everything but $f_{\theta}$ is discarded at the end of training.}
  \label{fig:04}
\end{figure*}

The other three embeddings are combined in the horizontal dimension to enrich the basic embeddings.
Compared with previous works that exploit image-text pairs~\cite{radford2021learning}, the layout information is a unique characteristic of VRDs.
Two layout embeddings are used in DocReL, namely the 1D layout embedding and the 2D layout embedding.
The 2D layout embedding exploits the bounding box of entities as the input, thus multiple text-token embeddings in one entity also correspond to the same 2D layout embedding.
To further distinguish different text tokens, a simple 1D sequential embedding ranges from 1 to maximum sequence length is adopted.
Besides, segment embedding and modality embedding are used to explicitly indicate different entities and different modalities.
Finally, the aggregated embedding is fed to the following Transformer module for feature fusion. 

\subsection{Relational Consistency Modeling}
\label{sec:pretrain-task}

Without loss of generality, we employ the framework of BYOL~\cite{grill2020bootstrap} as the basic contrastive learning paradigm for pre-training.
Specifically, for each training sequence $x$, we utilize two sets of augmentation strategies $t_{1}$ and $t_{2}$ to generate two augmented views $v_{1} = t_{1}(x)$ and $v_{2} = t_{2}(x)$. 
These augmented sequences~$(v_{1}, v_{2})$ are fed into the online feature extraction network~$f_{\theta}$ and the target network~$f_{\xi}$ to obtain aggregated multi-modal features of entities~$m_{v_{1}} = f_{\theta}(v_{1})$ and $m_{v_{2}} = f_{\xi}(v_{2})$, where $m_{v_{1}}, m_{v_{2}} \in \mathbb{R}^{N \times d}$, $N$ denotes the number of entities and $d$ means the dimension of features.
Different from previous methods~\cite{chen2020simple,chen2020big,he2020momentum,chen2020improved,chen2021empirical} that minimize the distance between the raw image representations of augmented $v_{1}$ and $v_{2}$, we design and harness two modules, namely Local Relational Consistency Modeling~(LRCM) and Global Relational Consistency Modeling~(GRCM), to minimize the distance of relational representations between different positive views.

\noindent\textbf{Local Relational Consistency Modeling: }
Entities in VRDs have local relational consistency.
That is, in different positive augmentation views, the original status between a pair of entities should not be modified, regardless of whether the relation exists or what the relation is.
To obtain the relational representations from entity features, we apply a relation extraction module to capture local relational context information. 
Specifically, multi-modal features of two paired entities are concatenated along the channel axis and then aggregated to the relational representations by the relation extraction module:
\begin{equation}
	\mathcal{R}^{{L}}(i,j) = f(\pi(i)\oplus\pi(j)), i,j \in \{1,2,..., N\},
	\label{eqn:01}
\end{equation}
where $\oplus$ indicates concatenation, $f(\cdot)$ is the function of relation aggregation implemented by a group of FC layers,  $\pi \in \mathbb{R}^{N \times d}$ are features of entities, $\mathcal{R}^{{L}} \in \mathbb{R}^{N^{2} \times d_{{L}}}$ means the extracted relational representations with the dimension of $d_{{L}}$.

In addition, within the online network, using the relational representations $\mathcal{R}^{{L}}_{v_{1}}$ of $v_{1}$  as the input, we conduct the projection operation $p_{v_{1}}^{L}=g^{L}_{\theta}(\mathcal{R}^{{L}}_{v_{1}})$ and the prediction operation $q^{L}_{\theta}(p_{v_{1}}^{L})$.
At the same time, the target projection $p^{L}_{v_{2}}$ of $v_{2}$ is output by the projection $g^{L}_{\xi}(\mathcal{R}^{{L}}_{v_{2}})$ in the target network.
Our local relational consistency loss is then defined as a mean squared error between the online network prediction and the target projection:
\begin{equation}
	\mathcal{L}_\text{LRCM} = \frac{1}{N^{2}}\sum^{N}_{i=1}\sum^{N}_{j=1}|| q^{L}_{\theta}(p^{L}_{v_{1}}) -  p^{L}_{v_{2}}|| \cdot M^{L},
\end{equation}
Note that the number of entities in different inputs is variable, so we pad them to the maximum size in each batch for parallel training and introduce a masked matrix $M^{L}\in \mathbb{R}^{N^{2} \times d_{{L}}}$ for masking the paddings.

\noindent\textbf{Global Relational Consistency Modeling: }
Local relational consistency focuses on the relation between a pair of entities.
However, there usually exist multiple entities in a single VRD, and their connections are complex and implicit.
Based on this observation, we take the global relational consistency into consideration.
The intuition is that relational structure built by all entities should be stable across different positive views. 
To calculate the global relations of a particular entity, we adopt a softmax layer to measure the similarities between this query entity and all entities, and finally produces its relational distribution. 
More concretely, given the query embedding of $i$-th entity output by feature extraction system is $\pi{(i)}$ and the relational entity is $\pi{(j)}$, the relational distribution is defined as:
\begin{equation}
	c(i, j) = 
	\frac{{\rm exp}(\pi{(i)}^{T}\cdot\pi{(j)}/\tau)}{\sum_{k=1}^{N}{\rm exp}(\pi{(i)}^{T}\cdot\pi{(k)}/\tau)}, i,j \in \{1,2,..., N\},
\label{eqn:globaldistribution}
\end{equation}
where $\tau$ denotes the temperature hyper-parameter.
Hence, the global relational distribution $\mathcal{R}^{{G}} \in \mathbb{R}^{N \times d_{G}}$ is generated by a concatenation operation:
\begin{equation}
	\mathcal{R}^{{G}}(i) = c({i},{1}) \oplus c({i},{2}) \oplus ... \oplus c({i},{N}), i \in \{1,2,..., N\}.
\end{equation}
Furthermore, the global relational consistency optimization is formulated as:
\begin{equation}
	\mathcal{L}_\text{GRCM} = \frac{1}{N}\sum_{i=1}^{N}|| q^{G}_{\theta}(g^{G}_{\theta}({\mathcal{R}^{{G}}_{v_{1}}(i)})) -  g^{G}_{\xi}({\mathcal{R}^{{G}}_{v_{2}}(i)})|| \cdot M^{G},
\end{equation}
where $\mathcal{R}^{{G}}_{v_{1}}$ and $\mathcal{R}^{{G}}_{v_{2}}$ are global relational distributions of $v_{1}$ and $v_{2}$, respectively.
$g^{G}$, $q^{G}$ and $M^{G}\in \mathbb{R}^{N \times d_{{G}}}$ are projection operation, prediction operation and masking matrix similar to that in $\mathcal{L}_\text{LRCM}$.
In view of both the local relational consistency and the global relational consistency, the overall loss of RCM is formulated as:
\begin{equation}
	\mathcal{L}_\text{RCM} = \mathcal{L}_\text{LRCM} + \mathcal{L}_\text{GRCM},
\end{equation}
and each loss coefficient is equally weighted. 
A more detailed description of RCM is given in supplementary material.

\subsection{Relational Feature Instantiation}
\label{sec:downstream}

RCM provides relational representations to promote downstream VRDs understanding tasks.
Among them, typical tasks are table structure recognition, key information extraction and reading order detection, \etc
Significant works have been done to fulfill these tasks, and a common solution is to exploit a sequence-to-sequence decoder to predict token-level target~\cite{wang2021layoutreader,zhang2020trie}.
As our relational representations are characterized by connections among semantic entities, such a trial will break the existing structures of representations and make them less effective. 

To further expand the effectiveness of our relational representations, we conduct surveys to investigate the common points of mainstream tasks, and try to solve them with a unified paradigm.
We observe that despite missions varying, their goals could be achieved by repeatedly classifying the binary relation between pairwise entities.
More concretely, considering two cells in the table structure recognition task, their pairwise relation is ``in the same column / row'' or not.
Considering two entities in the key information extraction task, they have pairwise relation ``key and value'' or not.
Moreover, considering two sentences in the reading order detection task, their pairwise relation is ``the former is ahead of the latter'' or on the contrary.
As the consequence, by repeatedly comparing pairwise entities by~\cref{eqn:01}, an intermediate relation matrix is generated, wherein each boolean element indicates if the specific relation exists or not.
Hence, downstream tasks could further extract their desired outputs via this relation matrix, which is an accessible instantiation of the relational representations.
With this paradigm, the relational representations could directly come into effect in downstream tasks.
Besides, we also observe that this unified paradigm yields better efficiency in particular downstream tasks, as demonstrated in~\cref{exp:exp-sota}.

\section{Experiments}
\label{sec:experiment}

\subsection{Datasets and Evaluation Protocol}
\label{sec:dataset}

In this section, we introduce the datasets for pre-training and evaluation.
Two datasets, RVL-CDIP~\cite{wei2020robust} and DocBank~\cite{li2020docbank} are utilized for pre-training.
We conduct evaluations on three downstream tasks:
1) Reading order detection task on ReadingBank~\cite{wang2021layoutreader},
2) Table structure recognition task on SciTSR~\cite{chi2019complicated}, ICDAR-2013~\cite{gobel2013icdar} and ICDAR-2019~\cite{gao2019icdar}, in which we follow the Setup-B setting in ~\cite{liu2021neural} where input by image along with layouts and contents,
3) Key information extraction task on FUNSD~\cite{jaume2019funsd} and CORD~\cite{park2019cord}, in which we focus on their entity linking tasks that rely on analyzing pairwise relation between entities.
We briefly introduce these datasets below.

\noindent\textbf{RVL-CDIP}~\cite{wei2020robust} consists of 400,000 gray-scale images in 16 classes, with 25,000 images per class. There are 320,000 training images, 40,000 validation images and 40,000 test images. 

\noindent\textbf{DocBank}~\cite{li2020docbank} contains 500K document pages with fine-grained token-level annotations for document layout analysis.
It is constructed using weak supervision from the free LaTeX documents.
It contains 400K for training, 50K for validation and 50K for testing.

\noindent\textbf{ReadingBank}~\cite{wang2021layoutreader} is a benchmark dataset for reading order detection built with weak supervision from WORD documents, which contains 500K document images with a wide range of document types as well as the corresponding reading order information.
This dataset is divided by ratio 8:1:1 for training, validation, and testing.

\noindent\textbf{SciTSR}~\cite{chi2019complicated} is a large-scale table structure recognition dataset, which contains 15,000 tables in PDF format and their corresponding structure labels obtained from LaTeX source files.
The official split assigns 12,000 for training and 3,000 for testing. 

\noindent\textbf{ICDAR-2013}~\cite{gobel2013icdar} is collected from digital documents, which contains 156 tables with annotation. 
Notice that this dataset does not provide training data. 
We randomly select $80\%$ of the data for training and the rest for testing by referring to a previous implementation~\cite{liu2021neural}.

\noindent\textbf{ICDAR-2019}~\cite{gao2019icdar} contains 600 tables for training and 150 tables for testing.
These tables are annotated with the ground truth of bounding boxes along with contents.

\noindent\textbf{FUNSD}~\cite{jaume2019funsd} consists of 199 real, fully annotated, scanned form images. 
The dataset is split into 149 training samples and 50 testing samples. 
Its official OCR annotation is utilized.

\noindent\textbf{CORD}~\cite{park2019cord} consists of receipts collected from shops and restaurants. 
It includes 800 receipts for training, 100 for validation, and 100 for testing with OCR annotations provided.

\noindent\textbf{Evaluation protocol.}
We employ the F1-score to evaluate the performance of the table structure recognition task and the key information extraction task.
For the reading order detection task, we select the BLEU score ~\cite{papineni2002bleu} that measures the n-gram overlaps between the prediction and ground truth as the evaluation metric.
We report the average BLEU in our experiments. 

\subsection{Implementation Details}
\label{sec:implementation}

\noindent\textbf{Experimental setting: }
We format the inputted VRD data for pre-training.
We select ResNet-50~\cite{he2016deep} as our visual feature extraction network and load its parameters from the official released pre-training model.
For datasets without text annotations, we apply the Tesseact engine~\cite{smith2007overview} to obtain the OCR results.
The document image is resized and padded to $512 \times 512$.
For feature embedding, the maximum length of the input sequence is limited to 512.
In addition, we exploit a 12-layer transformer encoder with 768 hidden sizes and 12 attention heads.
The initial weight of this transformer is loaded from the released LayoutLMv2~\cite{xu2020layoutlmv2} model.
The Adam optimizer is chosen for a warm-up with an initial $5 \times 10^{-5}$ learning rate.
In addition, we keep $1 \times 10^{-4}$ for 10 epochs and set a linear decay schedule for the rest of epochs. 
Based on Pytorch~\cite{paszke2017automatic}, we implement all benchmarks on a regular platform with 8 Nvidia V100 GPUs and 64GB memory.
We pre-train our DocReL architecture with DocBank dataset and RVL-CDIP dataset for 100 epochs with a batch size of 64.
For fine-tuning, we fine-tune the pre-trained model at all three downstream relational VRD understanding tasks for 50 epochs.
The batch size is 16 and the learning rate is $2 \times 10^{-5}$. 

\begin{table}[t]
    \newcommand{\tabincell}[2]{\begin{tabular}{@{}#1@{}}#2\end{tabular}}
    \centering
    \setlength{\tabcolsep}{1.0mm}

    \caption{Performance on table structure recognition tasks. 
    BROS and DocReL are pre-training frameworks with larger parameters.}
    \label{tab:01}
    \begin{tabular}{l|c|c|c|c}
        \toprule[1.5pt]
        \multirow{2}{*}{\tabincell{c}{Model}}
        &\multicolumn{1}{c|}{{SciTSR}} 
        &\multicolumn{1}{c|}{{ICDAR-13}}
        &\multicolumn{1}{c|}{{ICDAR-19}} 
        &\multirow{2}{*}{\tabincell{c}{Params}} \\
        \cline{2-4}
        & F1-score & F1-score & F1-score &  \\
        \hline
        DGCNN & 97.6 & 98.8 & -    & 0.8M       \\
        TabStruct & 99.1 & 99.2 & 96.6 & 4.7M       \\
        BROS & 99.5 & -    & -    & 113M       \\
        FLAG-Net & 99.6 & 99.3 & 96.2 & 1.9M       \\
        NCGM & 99.7 & 99.6 & 98.8 & 3.1M       \\
        \hline
        DocReL & \textbf{99.8}($\pm 0.1$) & \textbf{99.6}($\pm 0.2$) & \textbf{99.0}($\pm 0.2$) & 142M   \\
        \midrule[1.5pt] 
    \end{tabular}
    \vspace{-0.2cm}
\end{table}

\begin{table}[t]
    \newcommand{\tabincell}[2]{\begin{tabular}{@{}#1@{}}#2\end{tabular}}
    \centering
    \setlength{\tabcolsep}{3.2mm}
    \caption{Performance on key information extraction tasks.}
    \begin{tabular}{l|c|c|c}
        \toprule[1.5pt]
        \multirow{2}{*}{\tabincell{c}{Model}}  
        &\multicolumn{1}{c|}{{FUNSD}} 
        &\multicolumn{1}{c|}{{CORD}} 
        & \multirow{2}{*}{\tabincell{c}{Params}}   \\
        \cline{2-3}
        & F1-score & F1-score &  \\
        \hline
        SPADE & 41.3 & 92.5 & -\\
        StrucText & 44.1 & - & 166M  \\
        BERT & 27.7 & 92.8 & 110M  \\
        LayoutLM & 45.9 & 95.2 & 113M \\
        LayoutLMv2 & 42.9 & 95.6 & 200M \\
        \hline
        DocReL& \textbf{46.1}($\pm 0.2$) & \textbf{97.0}($\pm 0.1$) & 142M \\
        \midrule[1.5pt] 
    \end{tabular}
    \label{tab:02:el_sota}
\end{table}

\begin{table}[t]
    \newcommand{\tabincell}[2]{\begin{tabular}{@{}#1@{}}#2\end{tabular}}
    \centering
    \setlength{\tabcolsep}{2.7mm}
    \caption{Performance on the reading order detection task. 
    Thanks to the architecture of DocReL, its inference speed is an order of magnitude faster than the sequence-to-sequence counterpart, and this speed is consistent in different tasks.}
    \begin{tabular}{l|c|c|c}
        \toprule[1.5pt]
        \multirow{2}{*}{\tabincell{c}{Model}} & ReadingBank & \multirow{2}{*}{\tabincell{c}{Speed}} & \multirow{2}{*}{\tabincell{c}{Params}} \\
        \cline{2-2}
        & BLEU & & \\
        \hline
        Heuristic Method & 69.7 & - & - \\
        LayoutReader & 98.2 & 6.0s & 113M \\
        \hline
        DocReL & \textbf{98.4}($\pm 0.1$) & \textbf{0.2s} & 142M \\
        \midrule[1.5pt] 
    \end{tabular}
    \label{tab:03}
\end{table}

\noindent\textbf{Augmentation strategies of RCM: }
RCM relies on the positive views of instance for learning.
As introduced in~\cref{sec:approach}, entities in VRDs have complex relations, thus careful augmentations are required to keep the original local pairwise relations and the global relational distributions unchanged.

The basic operation unit is a single entity that has text, image and layout information, respectively.
Notice that these modalities are connected, that is, modification of one modality will influence others simultaneously.
Hence, these modalities could not be augmented separately.
As the consequence, we carefully design the augmentation strategy for RCM, which consists of two operations:
1) visual changes without text or layout modification.
In this case, only the appearance transformation of the image is conducted, which includes color distortion such as color brightness, contrast, saturation, hue, and Gaussian blur.
2) visual and layout changes without modification on text.
In addition to the previous operation, the width or height of the entity's bounding box is selected and is resized with ratio [0.85, 1.15].
The validity of this operation is checked afterward to avoid the overlap of entities.
The visual features are reshaped correspondingly.
These two operations are randomly selected for augmentation during training.
To validate the effectiveness of this approach, we generate $10$ augmented views of $1, 000$ VRDs and conduct the manual check.
We observe that only $5$ VRDs encounter slight relation modifications such as table cell changing its row or col.
Hence, this augmentation module is sufficient to produce robust positive views for training.

\noindent\textbf{Post-processing for downstream tasks:}
For fair comparisons on various relational understanding tasks, we perform light-weighted post-processing on the results of DocReL. 
A more detailed description is given in supplementary material.

\subsection{Comparison with SOTA}
\label{exp:exp-sota}

\begin{table}[t]
    \newcommand{\tabincell}[2]{\begin{tabular}{@{}#1@{}}#2\end{tabular}}
    \centering
    \small
    \setlength{\tabcolsep}{0.2mm}

    \caption{Ablation studies of pre-training tasks on SciTSR, FUNSD and ReadingBank datasets. 
    The models with ``*'' denote using the same datasets for pre-training and the same relational prediction module for downstream tasks consistent with ``DocReL''. 
    ``BYOL$^{\dag}$'' means that the multi-modal features of entities serve as the supervisory signal of consistency instead of original image features.}

    \begin{tabular}{l|l|c|c|c}
        \toprule[1.5pt]
        \multirow{2}{*}{Model} 
        & \multirow{2}{*}{\tabincell{c}{Pre-training Tasks}}  
        & \multicolumn{1}{c|}{{SciTSR}} 
        & \multicolumn{1}{c|}{{FUNSD}}
        & \multicolumn{1}{c}{{ReadingBank}}  \\
        \cline{3-5}
        &  & F1-score & F1-score & BLEU \\
        \hline
        \multirow{4}{*}{LayoutLMv2*} 
        &MVLM                       & 97.80 & 40.47 & 96.24 \\
        &MVLM+BYOL$^{\dag}$         & 98.04 & 40.96 & 96.38 \\
        &MVLM+LRCM         & 98.15 & 42.03 & 96.74 \\
        &MVLM+LRCM+GRCM    & \textbf{98.45} & \textbf{42.78} & \textbf{97.08} \\
        \hline
        \multirow{4}{*}{StrucText*} 
        &MVLM                       & 98.65 & 42.86 & 97.38 \\
        &MVLM+BYOL$^{\dag}$         & 98.76 & 43.29 & 97.52 \\
        &MVLM+LRCM         & 99.07 & 44.33 & 97.88 \\
        &MVLM+LRCM+GRCM    & \textbf{99.36} & \textbf{45.80} & \textbf{98.11} \\
        \hline
        \multirow{4}{*}{DocReL} 
        &MVLM                       & 98.74 & 42.74 & 97.46 \\
        &MVLM+BYOL$^{\dag}$         & 98.86 & 43.53 & 97.62 \\
        &MVLM+LRCM         & 99.31 & 44.62 & 98.04 \\
        &MVLM+LRCM+GRCM    & \textbf{99.78} & \textbf{46.08} & \textbf{98.41} \\
        \midrule[1.5pt] 
    \end{tabular}
    \label{tab:02}
\end{table}

\noindent\textbf{Table structure recognition: }
The comparison results are shown in~\cref{tab:01}. 
Specifically, our method obtains an F1-score of $99.8\%$ in SciTSR, and surpasses that of NCGM~\cite{liu2021neural} by $0.1\%$. 
We obtain an F1-score of $99.6\%$ in ICDAR-2013 and $99.0\%$ in ICDAR-2019.
To make sure the performance gain is statistically stable, we repeat our experiments five times to avoid random fluctuations and attach the standard deviation to the scores.
The following experiments also adopt this method.

\begin{figure}[t]
    \centering
    \includegraphics[width=1\linewidth]{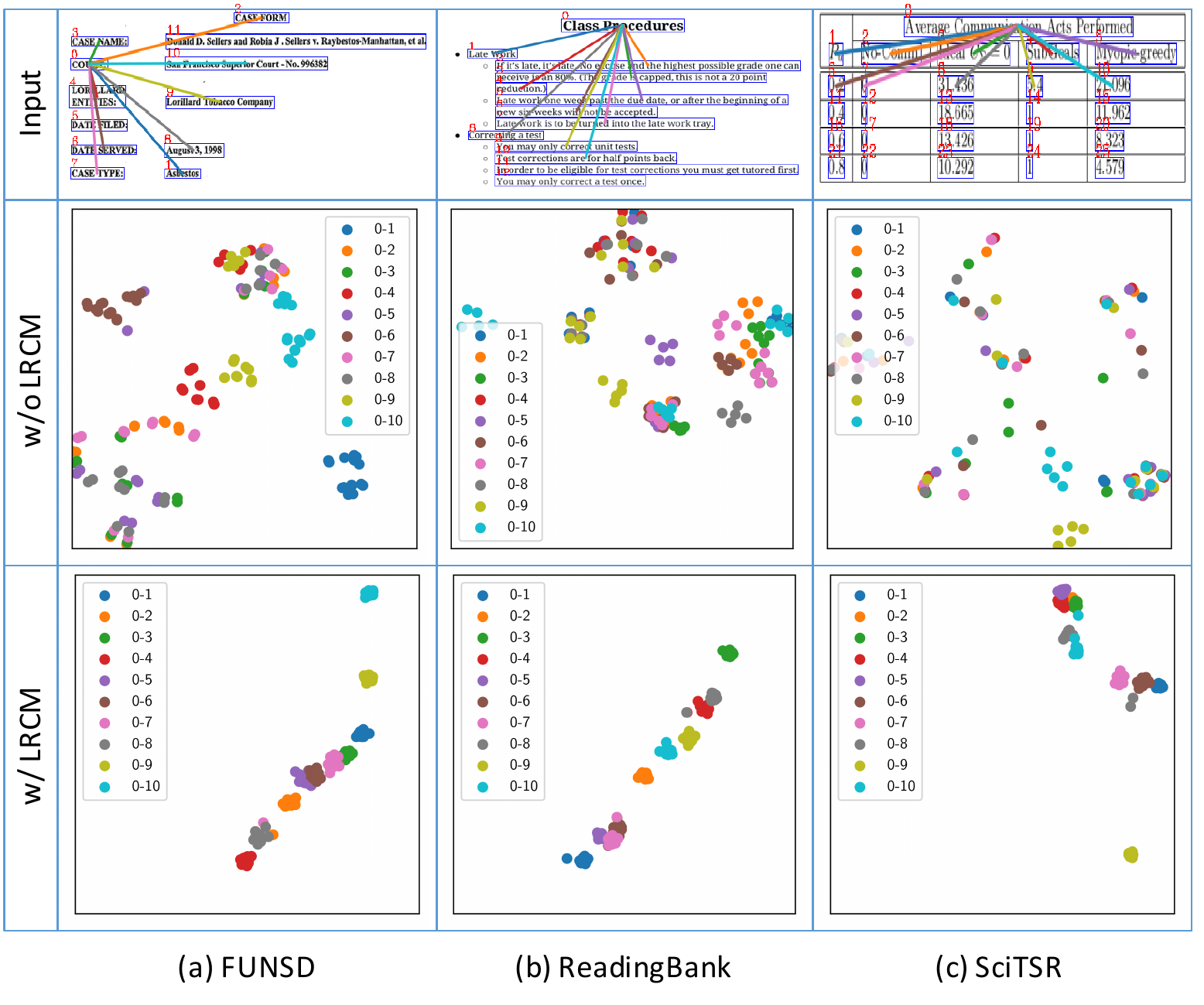}
    \caption{Effectiveness of LRCM.
    We visualize the local relational features on FUNSD, ReadingBank and SciTSR datasets by the t-SNE tool. 
    Compared with ``w/o LRCM'', the relational representations across different views with LRCM show more consistency. 
    ``w/'' and ``w/o'' are short for ``with'' and ``without''.
    Best viewed in color and zoom in.}
    \label{fig:local}
\end{figure}

\begin{figure*}[t]
    \centering
    \includegraphics[width=1\textwidth]{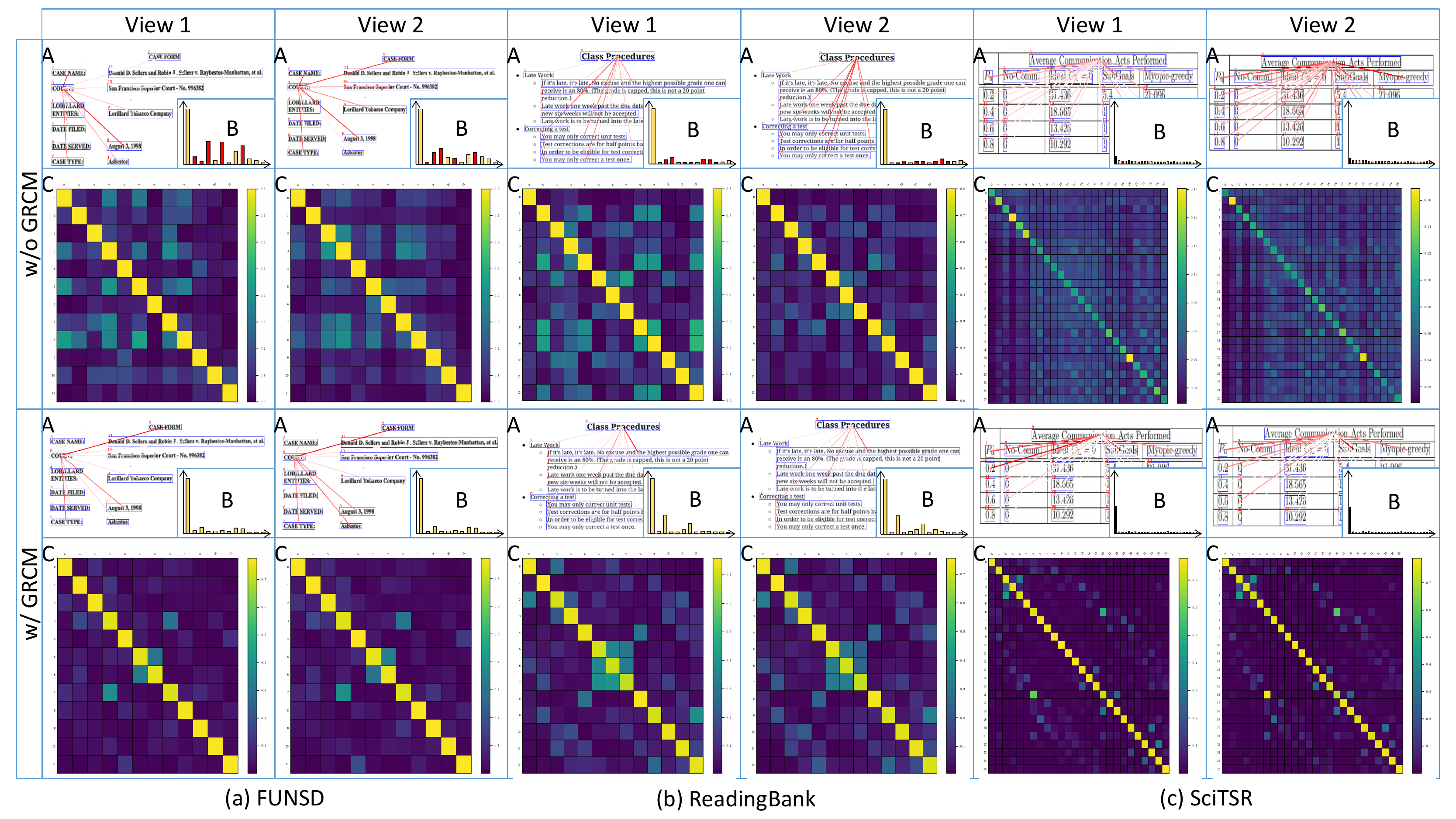}
    \caption{Effectiveness of GRCM.
    GRCM learns the relation distribution formed by all entities (part ``A''), and we particularly display connections of the $0$-th entity for clarification.
    A stronger red intensity indicates the stronger connections between pairwise entities, and the histogram (part ``B'') visualizes its connections from another perspective.
    With the help of GRCM, the global relation distributions (part ``C'') are more stable in different positive views.
    Best viewed in color and zoom in.}
    \label{fig:global}
\end{figure*}

\noindent\textbf{Key information extraction: }
The comparison results are shown in~\cref{tab:02:el_sota},  
our method achieves an F1-score of $46.1\%$ in FUNSD. 
Compared with the previous work that exploits a customized paired box direction~\cite{li2021structext} task for key-value pairing,  our proposed RCM is a more general relational learning task.
Besides, our method achieves an F1-score of $97.0\%$ in CORD and surpasses the suboptimal by $1.4\%$.

\noindent\textbf{Reading order detection: }
The comparison results are shown in~\cref{tab:03}.
The reading order detection task is an emerging task that firstly proposed by LayoutReader~\cite{wang2021layoutreader}.
DocReL achieves an F1-score $98.4\%$ which surpasses LayoutReader by $0.2\%$.
Notice that LayoutReader exploits a sequence-to-sequence decoder, and its inference speed increases quadratically according to the input sequence length.
With the same maximum sequence length $512$, the average inference speed of DocReL is an order of magnitude faster.
This advantage owes to the system architecture of DocReL.

A more detailed analysis on the above tasks is given in supplementary material.

\subsection{Ablation Study}
We perform analytic experiments to investigate the effectiveness of RCM, in both view of local relational consistency and global relational consistency.
To explore the accessibility of RCM, we carry out experiments using several frameworks~(\textit{i.e.}, LayoutLMv2~\cite{xu2020layoutlmv2}, StrucText~\cite{li2021structext} and DocReL) as the backbones.
For a fair comparison, we utilize the same datasets introduced in~\cref{sec:dataset} for pre-training and employ the same relational feature instantiation module in~\cref{sec:downstream} for downstream tasks. 
Notice that some approaches are not designed to provide explicit entity-level features.
In StrucText, the features of entities are fetched from visual tokens.
In LayoutLMv2, we calculate features of each entity by averaging its text tokens.

\noindent\textbf{Effectiveness of the LRCM: }
We validate the effectiveness of LRCM and report the result in~\cref{tab:02}.
We observe that applying contrastive learning ~(``MVLM+BYOL'' and ``MVLM+LRCM'') achieves better results than the baseline method~(``MVLM''). 
Moreover, LRCM increases 0.42\% BLEU on ReadingBank, 0.45\% F1-score on SciTSR and 1.09\% on FUNSD compared with BYOL. 
And introducing LRCM shows consistent trends across the frameworks of LayoutLMv2 and StrucText.
This result shows that contrastive learning on pairwise relational representations is more effective, compared with raw contextual features.

\noindent\textbf{Effectiveness of the GRCM:}
We also compare the performance with and without GRCM equipped on each model. 
As demonstrated in~\cref{tab:02}, we observe that GRCM boosts the performance with at least 0.24\%, 0.3\%, and 0.75\% improvement on ReadingBank, SciTSR, and FUNSD across different frameworks.
These results prove the effectiveness of GRCM.

\subsection{Further Analysis on RCM}

\noindent\textbf{What does RCM learn from local relational consistency?}
To provide intuitive examples for explaining the mechanism of RCM, we leverage the t-SNE~\cite{van2008visualizing} to visualize the relational representations learned by DocReL.
For each input VRD image, we conduct augmentation to obtain $15$ different views, then send them to 1) DocReL with LRCM + MVLM and 2) DocReL with only MVLM task.
As shown in~\cref{fig:local}, each point in the t-SNE visualizations stands for the relational representations between the $i$-th entity and the $j$-th entity, and the colors are used to distinguish different pairs. 
For clear illustration, we only visualize $10$ pairs of them across multiple views.
We observe that the representations generated by LRCM~(w/ LRCM) tend to maintain consistency, which means LRCM pulls entity pairs with similar relations closer.
Once removing LRCM, the relational representations obtained by pure MVLM~(w/o LRCM) are still implicit.
Using \cref{fig:local}~(a) as the example, the Key-Value relation~(light blue color) between the 0-th entity~(``COURT:'') and the 10-th entity~(``San Fran.'') has more obvious distinctions compared with other unpaired relations in the t-SNE result w/ LRCM.
These observations demonstrate that LRCM is able to guide models to understand pairwise relations comprehensively, which is especially useful for relation understanding tasks.

\noindent\textbf{What does RCM learn from global relational consistency?} 
To further explore the effectiveness of global relational consistency, we visualize the global relational distribution formed by all entities by~\cref{eqn:globaldistribution}, as shown in~\cref{fig:global}. 
Compared with the result of MVLM~(w/o GRCM), we observe that GRCM~(w/ GRCM) learns more stable global relation distributions under different positive views.
We take the 0-th entity~(``COURT:'') and the 2-th entity~(``CASE FORM'') in the FUNSD case (\cref{fig:global}~(a)) as the example.
With only pure MVLM task, the connection between them is not stable in two different views.
On the other hand, with the assistance of GRCM, their relational connection rarely changes in different views.
Hence, GRCM provides global relational consistency to build better relational representations.
More visualization results about GRCM, especially for key information extraction and reading order detection tasks are given in supplementary material.

\section{Conclusion}
\label{sec:conclusion}
Relational understanding is critical for a number of VRDs understanding tasks.
To provide a more suitable feature representation for these tasks, we propose DocReL, a document relational representation learning framework.
DocReL harnesses a novel self-supervised pre-training task named Relational Consistency Modeling (RCM).
RCM exploits the fact that existing relations should be consistent in differently augmented positive views, thus is unrestricted to specific relational understanding tasks.
RCM provides relational representations which are more compatible to the urgent need of downstream tasks, even with zero-knowledge about the exact relational definition.
Experiment results show that DocReL achieves state-of-the-art results on a wide variety of downstream relational understanding tasks.

\appendix
\section{Appendix}

\subsection{Post-processing for Downstream Tasks}
With the relational feature instantiation module, an intermediate relation matrix is generated, wherein each boolean element indicates if the specific connection exists or not.
In addition, we employ a lightweight post-process to instantiate the relation matrix for each downstream task.

\noindent\textbf{Post-processing for table structure recognition: }
For each pair of entities, we judge whether they belong to the same row / column or not. 
Taking the row-relation between $i$-th and $j$-th entities as an example, if the predicted value in the relation matrix surpasses $0.5$, it indicates that they belong to the same row. 
Then we calculate an F1-score by matching linked pairwise indexes with ground truth. 
Similarly, the above operations are also applied to the column. 
The final result is computed by averaging row-wise and col-wise scores.

\noindent\textbf{Post-processing for key information extraction: }
To obtain the linked key-value pairwise results, a post-processing step similar to the one in table structure recognition is leveraged to match text contents for key information extraction. 
Specifically, for each predicted relation, we concatenate the strings of the start index~(key) and the end index~(value) to get the structural result.

\noindent\textbf{Post-processing for reading order detection: }
We follow the principle of ``the former is ahead of the latter'' to recover the reading order of the inputted VRD paragraphs. 
For all unsorted sentences, we compare their pairwise relation to determine the headmost entity, and put it in the sorted sequence. 
Then we repeat the above step until all entities are sorted. 
The BLEU score between the prediction and ground truth is calculated to measure the performance of reading order detection.

\subsection{Effectiveness of RCM on various datasets}
\begin{figure}[b]
	\centering
	\includegraphics[width=1\linewidth]{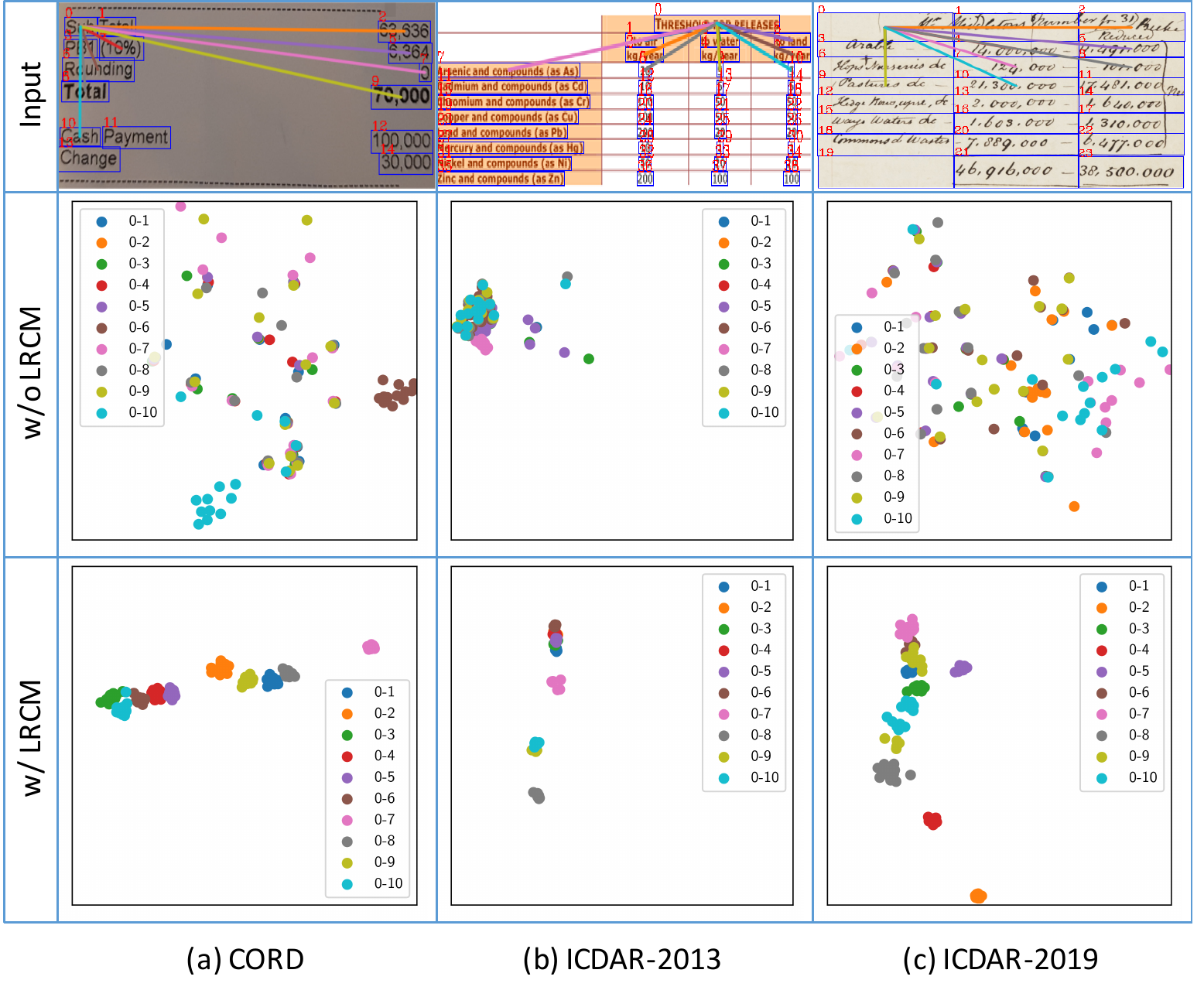}
	\caption{More local relational representations visualized with t-SNE sampled from CORD, ICDAR-2013 and ICDAR-2019 datasets. Best viewed in color and zoom in.}
	\label{fig:appendix1}
\end{figure}

\begin{figure}[b]
	\centering
	\includegraphics[width=.48\textwidth]{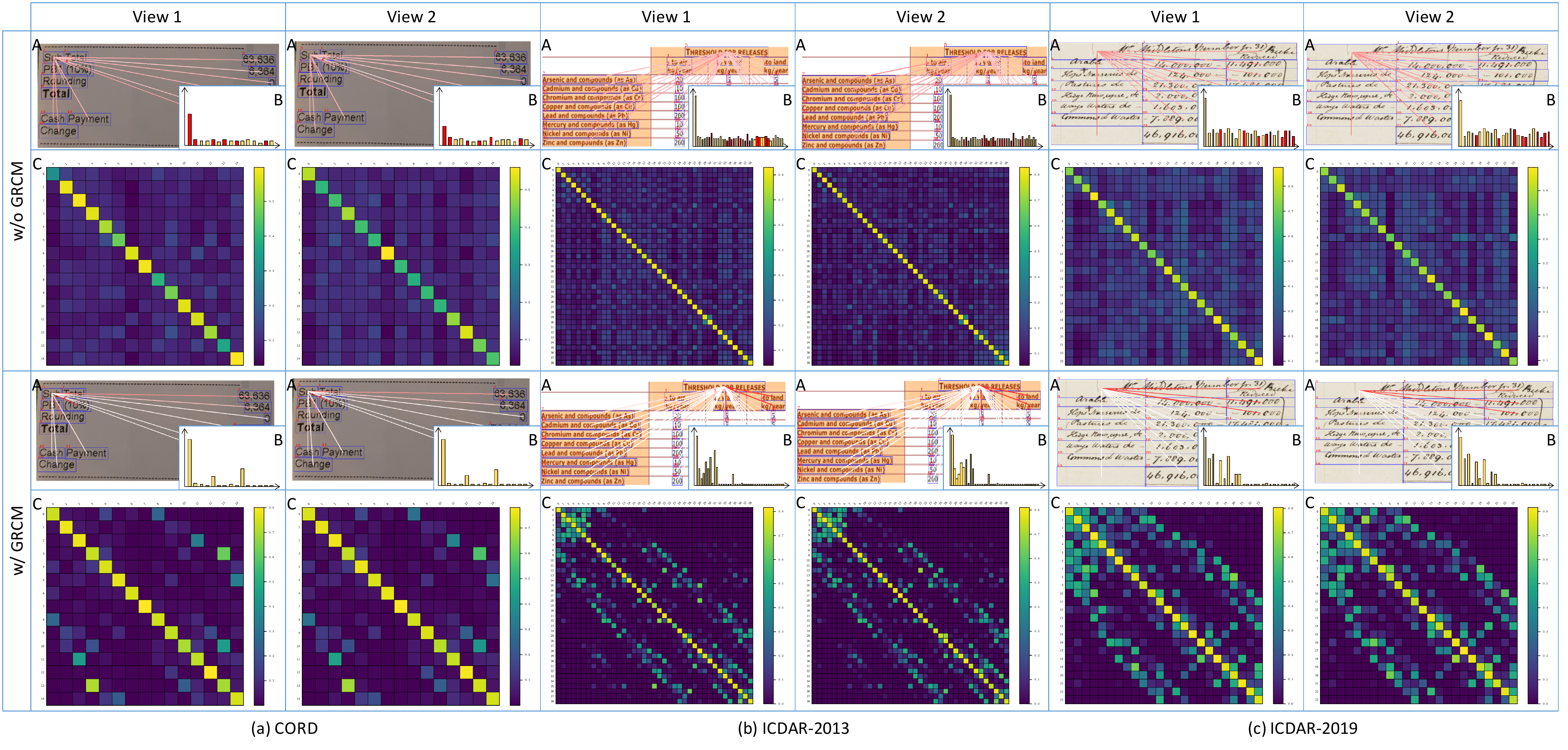}
	\caption{More visualizations of global relational distributions sampled from CORD, ICDAR-2013 and ICDAR-2019 datasets. Best viewed in color and zoom in.}
	\label{fig:appendix2}
\end{figure}

\cref{fig:appendix1} and \cref{fig:appendix2} demonstrate more visualizations of RCM on various benchmarks. 

\subsection{Effectiveness of DocReL on real-world datasets}
\begin{figure*}[t]
\centering

    \subfloat[FUNSD (Pure MVLM)]{\includegraphics[width=.33\textwidth]{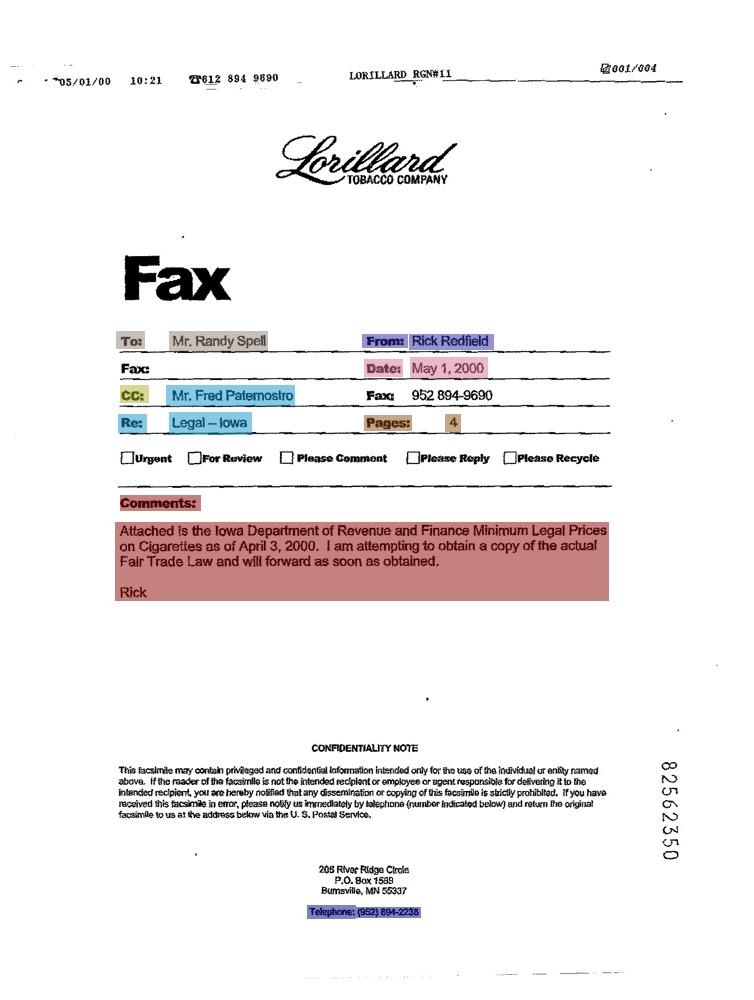}}
    \hfill
    \subfloat[FUNSD (DocReL)]{\includegraphics[width=.33\textwidth]{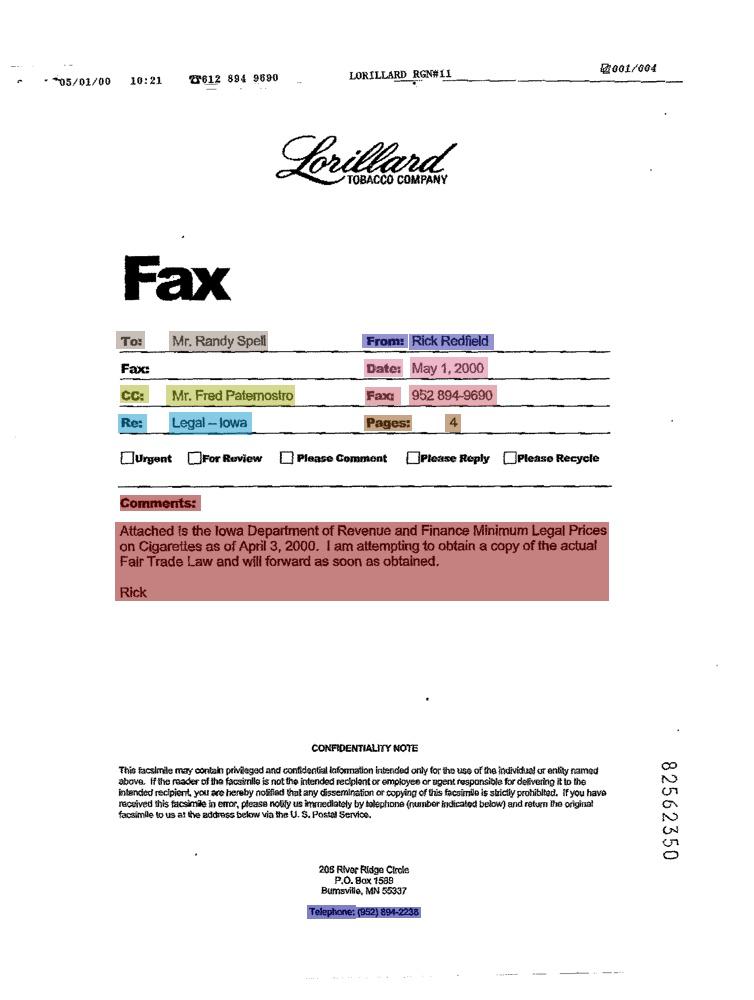}}
    \hfill
    \subfloat[FUNSD (Ground Truth)]{\includegraphics[width=.33\textwidth]{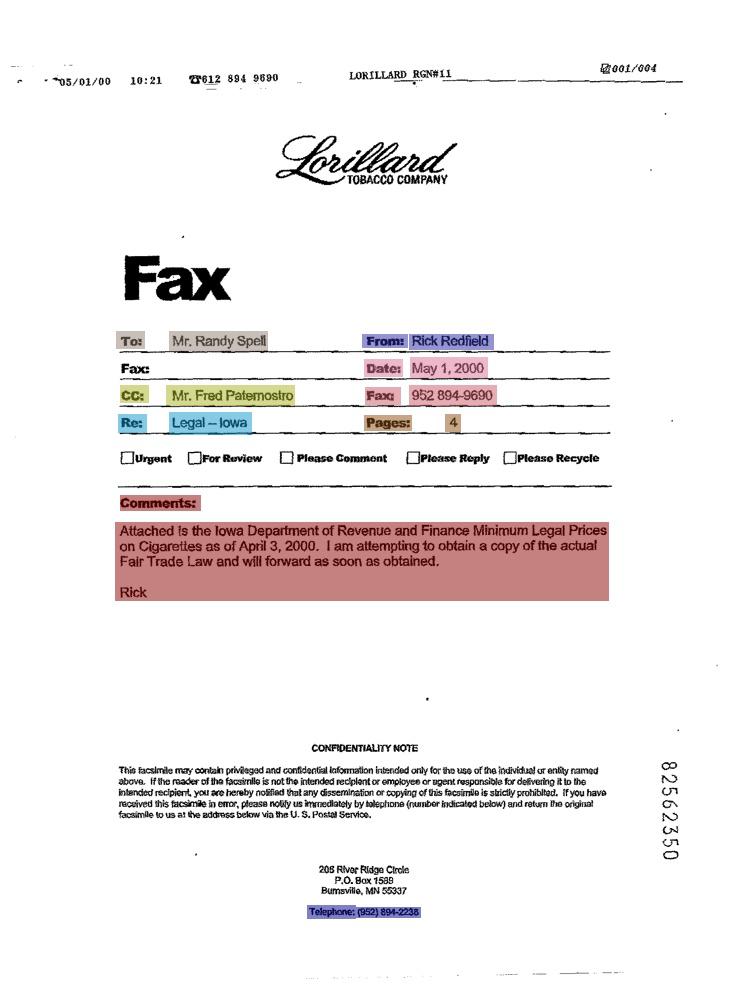}}

    \subfloat[CORD (Pure MVLM)]{\includegraphics[width=.33\textwidth]{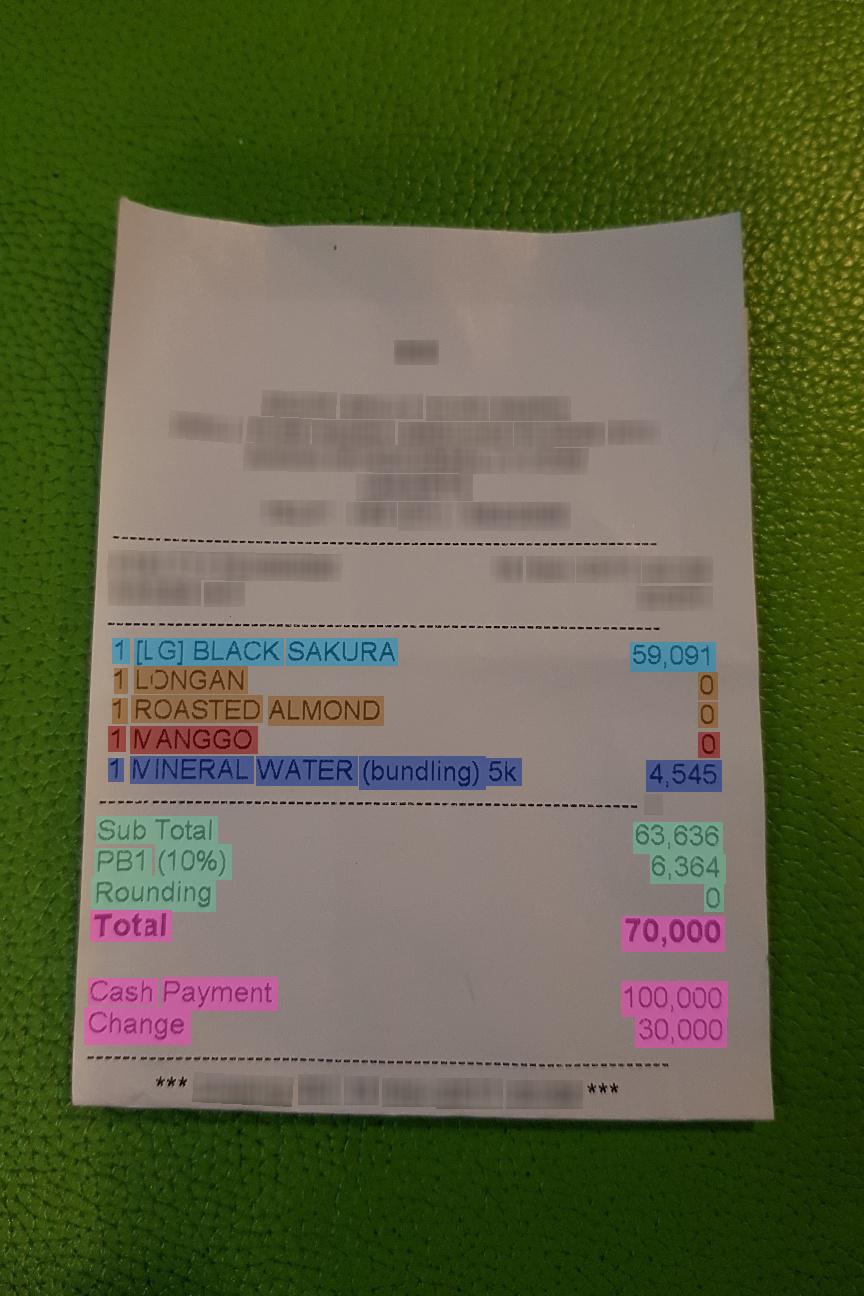}}
    \hfill
    \subfloat[CORD (DocReL)]{\includegraphics[width=.33\textwidth]{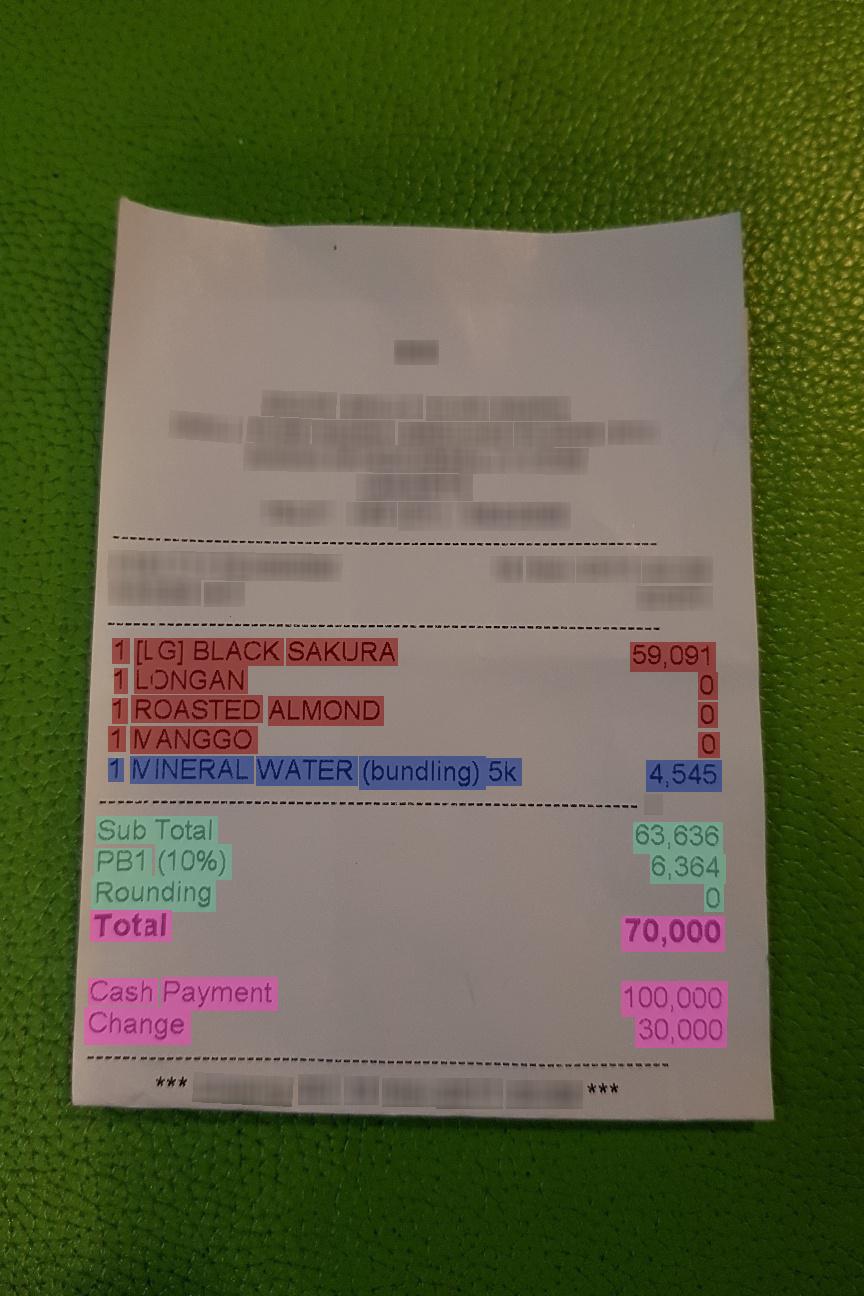}}
    \hfill
    \subfloat[CORD (Ground Truth)]{\includegraphics[width=.33\textwidth]{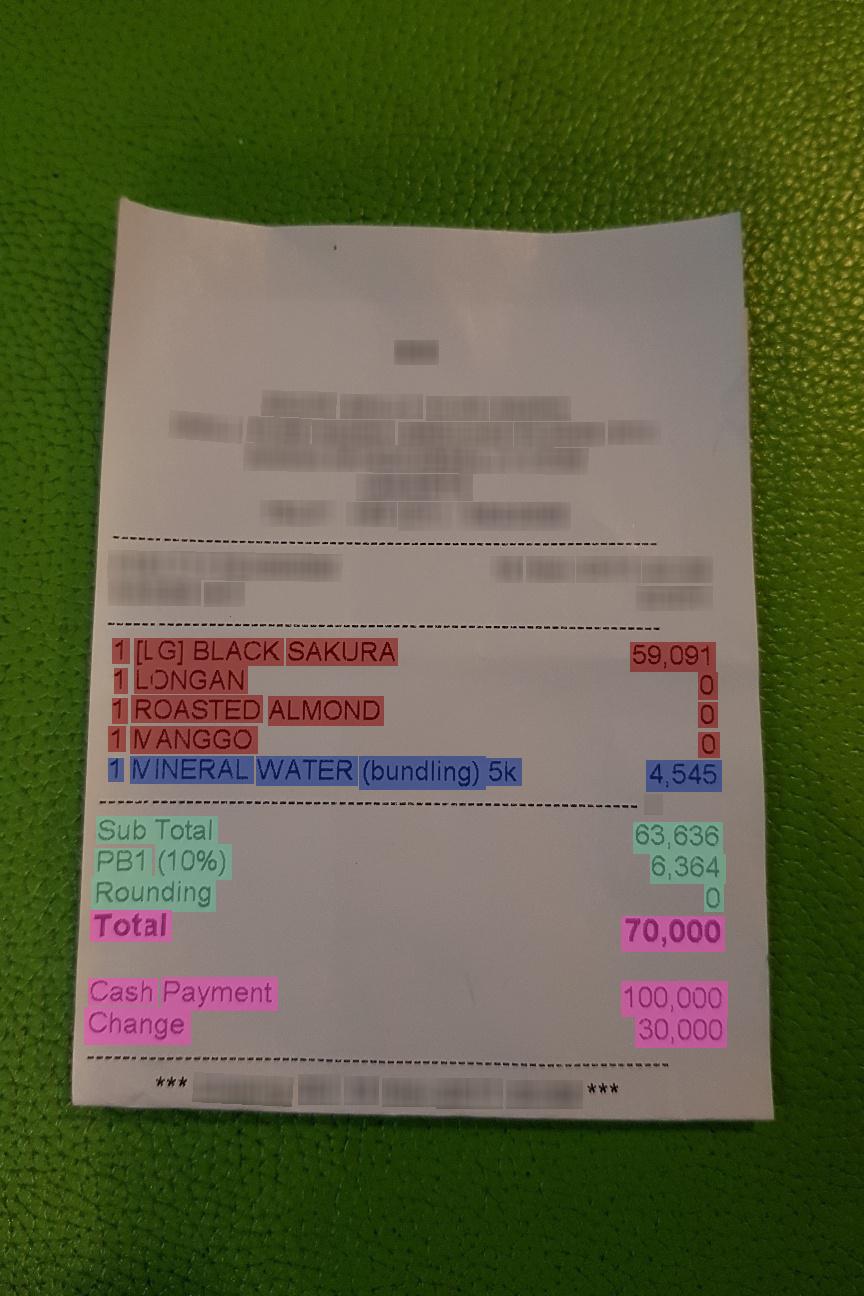}}

  \caption{Sample output of DocReL on Key Information Extraction datasets. 
  Items with the same color are key-value pairs in FUNSD or key-value groups in CORD.
  Best viewed in color and zoom in.}
  \label{fig:appendix31}
\end{figure*}

\begin{figure*}[t]
\centering
    \subfloat[SciTSR (Pure MVLM)]{\includegraphics[width=.33\textwidth]{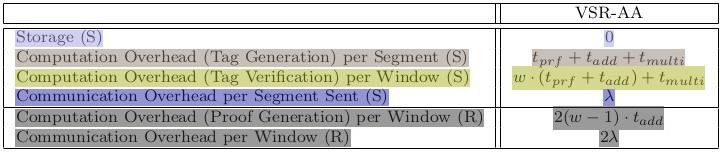}}
    \hfill
    \subfloat[SciTSR (DocReL)]{\includegraphics[width=.33\textwidth]{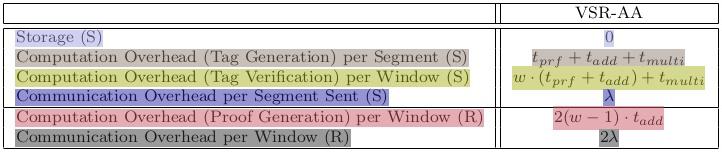}}
    \hfill
    \subfloat[SciTSR (Ground Truth)]{\includegraphics[width=.33\textwidth]{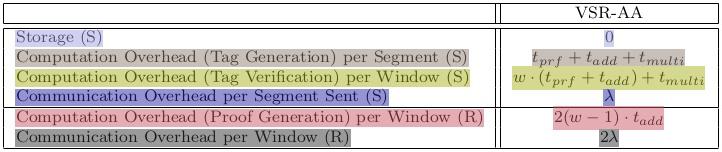}}

    \subfloat[ICDAR-13 (Pure MVLM)]{\includegraphics[width=.33\textwidth]{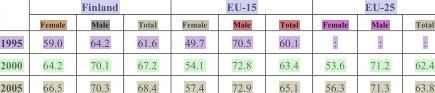}}
    \hfill
    \subfloat[ICDAR-13 (DocReL)]{\includegraphics[width=.33\textwidth]{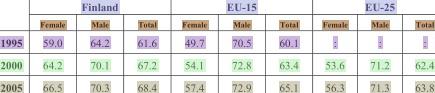}}
    \hfill
    \subfloat[ICDAR-13 (Ground Truth)]{\includegraphics[width=.33\textwidth]{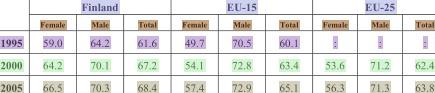}}

    \subfloat[ICDAR-19 (Pure MVLM)]{\includegraphics[width=.33\textwidth]{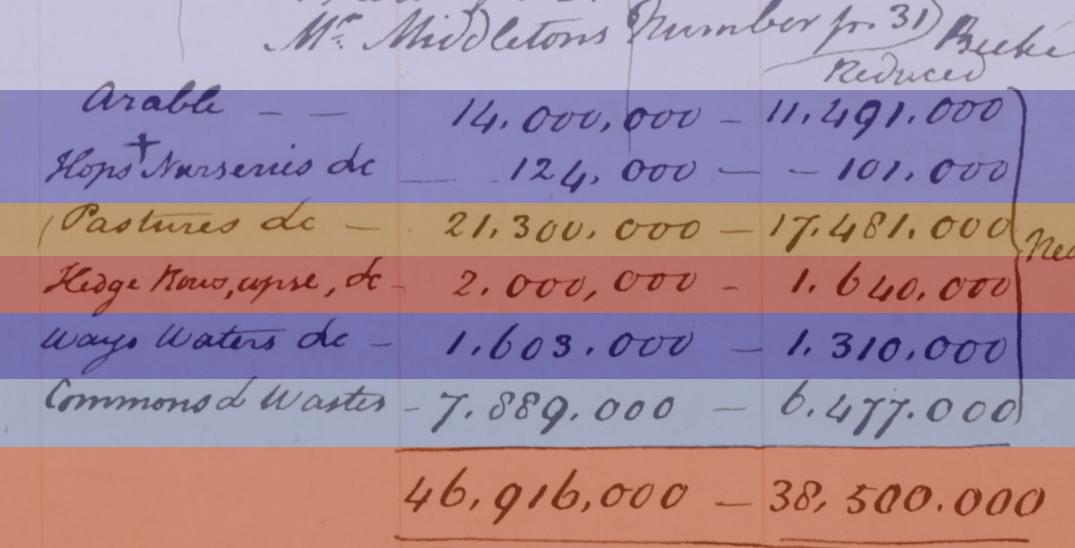}}
    \hfill
    \subfloat[ICDAR-19 (DocReL)]{\includegraphics[width=.33\textwidth]{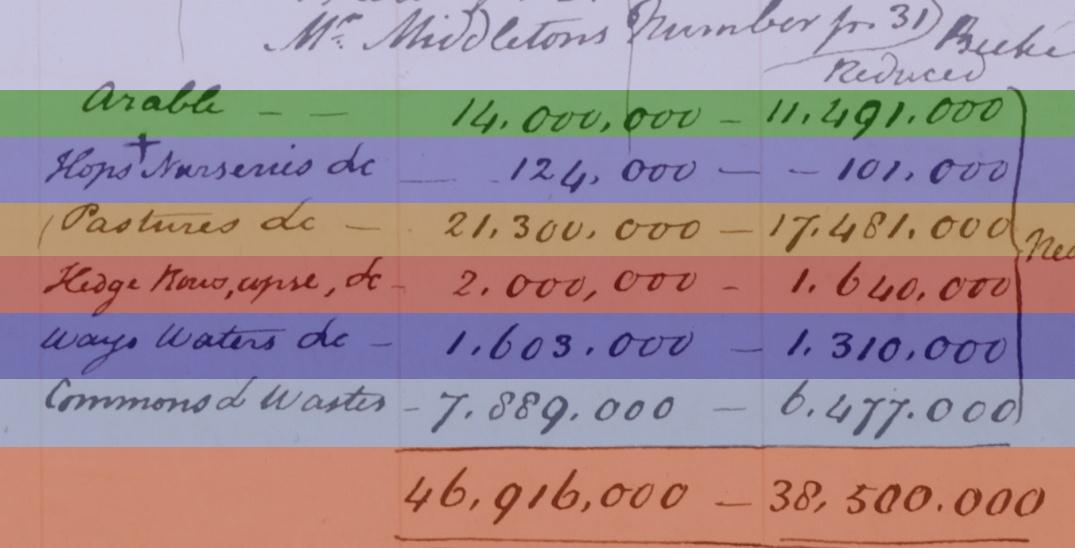}}
    \hfill
    \subfloat[ICDAR-19 (Ground Truth)]{\includegraphics[width=.33\textwidth]{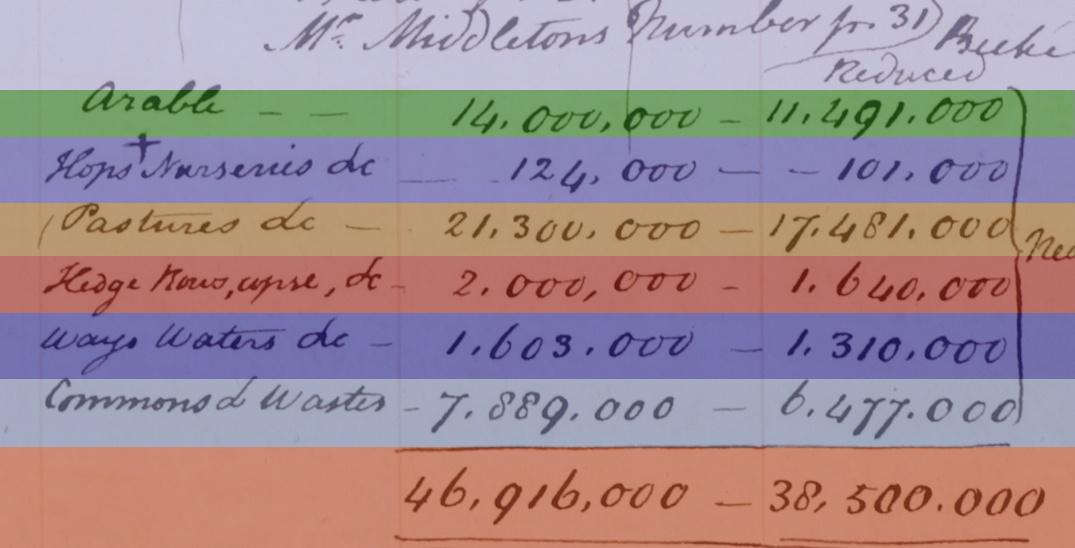}}

  \caption{Sample output of DocReL on Table Structure Recognition datasets. 
  Items with the same color are on the same row.
  Best viewed in color and zoom in.}
  \label{fig:appendix32}
\end{figure*}

\begin{figure*}[t]
\centering
    \subfloat[ReadingBank (Pure MVLM)]{\includegraphics[width=.33\textwidth]{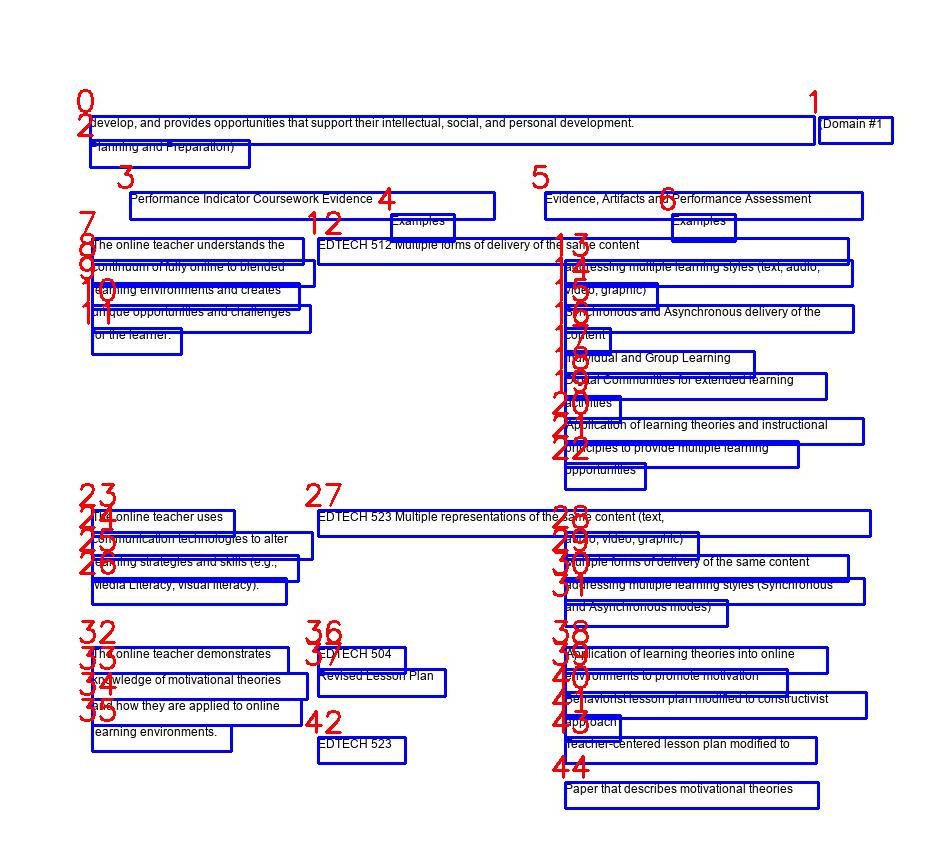}}
    \hfill
    \subfloat[ReadingBank (DocReL)]{\includegraphics[width=.33\textwidth]{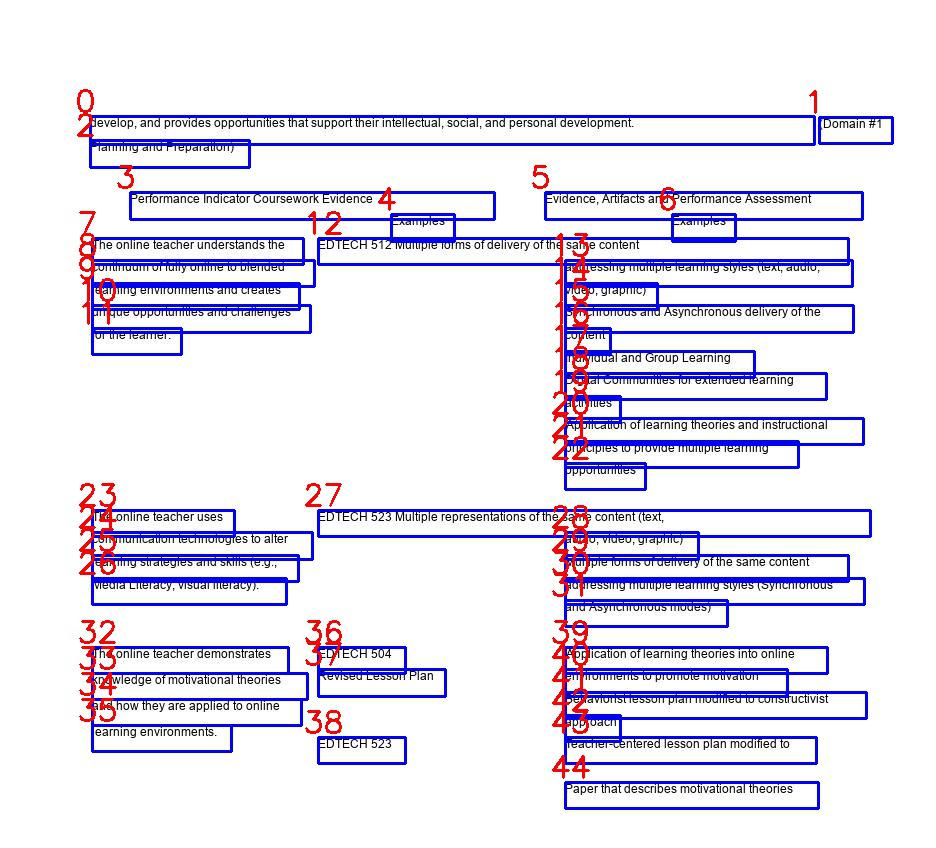}}
    \hfill
    \subfloat[ReadingBank (Ground Truth)]{\includegraphics[width=.33\textwidth]{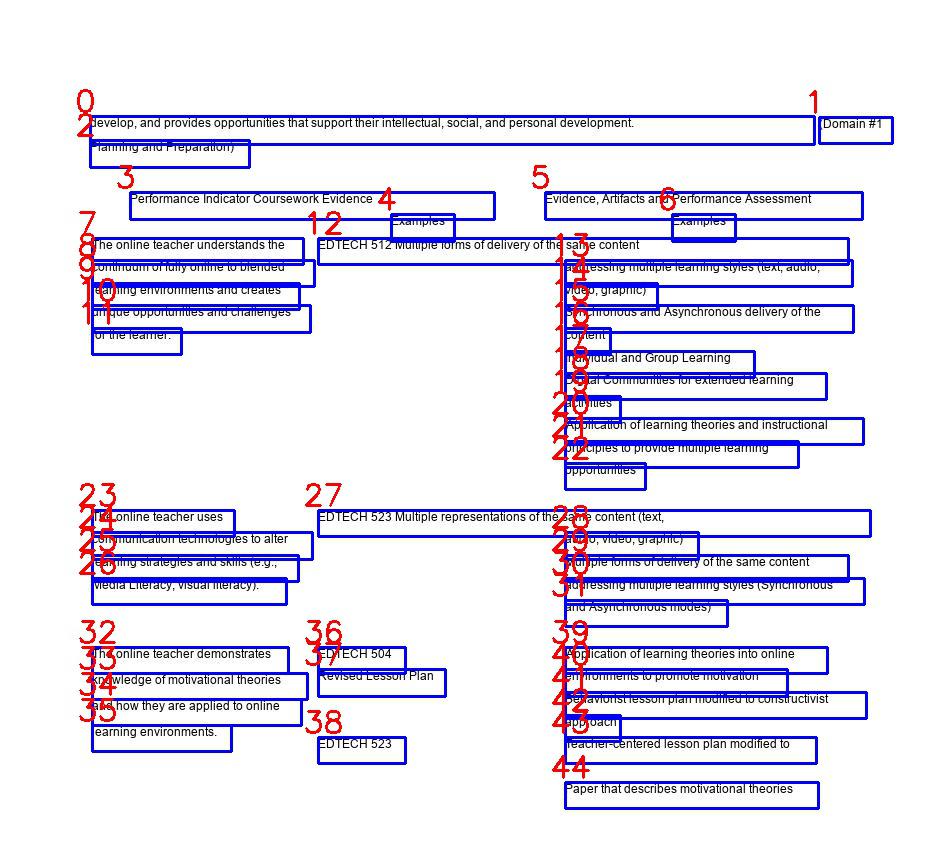}}

  \caption{Sample output of DocReL on the ReadingBank dataset. 
  Red digits on the top-left corners indicate the reading order of sentences.
  As ReadingBank is a pure-text dataset, we render the displayed images by using text annotations as the input with a default font characteristic.
  Best viewed in color and zoom in.}
  \label{fig:appendix33}
\end{figure*}

\cref{fig:appendix31}, \cref{fig:appendix32} and \cref{fig:appendix33} demonstrate the effectiveness of DocReL on real-world datasets of VRDs.
DocReL is able to learn better relational representations, which are expected to benefit downstream relational understanding tasks.

\subsection{Process of Relational Consistency Modeling (RCM)}
For clarity, the detailed process of RCM is shown in Alg.~\ref{alg}.

\begin{algorithm}[h]
	\small
	\newcommand{\tabincell}[2]{\begin{tabular}{@{}#1@{}}#2\end{tabular}}
	\SetKwInput{KwInput}{Input}      
	\SetKwInput{KwOutput}{Output}    
	\SetKwProg{Fn}{Function}{:}{}
	\DontPrintSemicolon
	\KwInput{
		\par
		\begin{tabular}{ll}
			$\mathcal{X}$, $\mathcal{T}$ & set of inputs and strategies of augmentation\\
			$\theta$, $f_{\theta}$, $g_{\theta}$ and $q_{\theta}$ & \tabincell{l}{initial online parameters, encoder, projector, \\ and predictor}\\
			$\xi$, $f_{\xi}$, $g_{\xi}$ & \tabincell{l}{initial target parameters, target encoder, and \\  target projector}\\
			$K$, $\tau$ & \tabincell{l}{total number of optimization steps and  \\  target network update schedule}\\
	\end{tabular}}
	
	\KwOutput{$f_{\theta}$}
	
	\tcc{Local Relational Representations.}
	\SetKwFunction{FLRR}{{LRR}}
	\Fn{\FLRR{$m$}}{
		\For{i $\leq$ N} {
			\For{j $\leq$ N} {
				$\pi(i),\pi(j) ~\leftarrow~ m[i], m[j]$ \;
				$\mathcal{R}^{{L}}(i,j) ~\leftarrow~ f(\pi(i)\oplus\pi(j))$ \;
			}
		}
		\KwRet $\mathcal{R}^{{L}}$
	}
	
	\tcc{Local Relational Consistency Modeling.}
	\SetKwFunction{FLRCM}{{LRCM}}
	\Fn{\FLRCM{$m_{v_{1}}, m_{v_{2}}$}}{
		$\mathcal{R}^{{L}}_{v_{1}}, \mathcal{R}^{{L}}_{v_{2}} ~\leftarrow~ \text{LRR}_{v_{1}}(m_{v_{1}}), \text{LRR}_{v_{2}}(m_{v_{2}})$ \;
		$p_{v_{1}}^{L},p_{v_{2}}^{L}~\leftarrow~g^{L}_{\theta}(\mathcal{R}^{{L}}_{v_{1}}),g^{L}_{\xi}(\mathcal{R}^{{L}}_{v_{2}})$ \;
		$\mathcal{L}_\text{LRCM} ~\leftarrow~ \frac{1}{N^{2}}\sum^{N}_{i=1}\sum^{N}_{j=1}|| q^{L}_{\theta}(p^{L}_{v_{1}}) -  p^{L}_{v_{2}}|| \cdot M^{L}$ \;
		\KwRet $\mathcal{L}_\text{LRCM}$
	}
	
	\tcc{Global Relational Distributions.}
	\SetKwFunction{FGRD}{{GRD}}
	\Fn{\FGRD{$m$}}{
		\For{i $\leq$ N} {
			\For{j $\leq$ N} {
				$\pi(i), \pi(j) ~\leftarrow~ m[i], m[j]$ \;
				$c(i, j) ~\leftarrow~ \frac{{\rm exp}(\pi{(i)}^{T}\cdot\pi{(j)}/\tau)}{\sum_{k=1}^{N}{\rm exp}(\pi{(i)}^{T}\cdot\pi{(k)}/\tau)}$ \;
			}
			$\mathcal{R}^{{G}}(i) ~\leftarrow~ c({i},{1}) \oplus c({i},{2}) \oplus ... \oplus c({i},{N})$ \;
		}
		\KwRet $\mathcal{R}^{{G}}$
	}
	
	\tcc{Global Relational Consistency Modeling.}
	\SetKwFunction{FGRCM}{{GRCM}}
	\Fn{\FGRCM{$m_{v_{1}}, m_{v_{2}}$}}{
		$\mathcal{R}^{{G}}_{v_{1}}, \mathcal{R}^{{G}}_{v_{2}} ~\leftarrow~ \text{GRD}_{v_{1}}(m_{v_{1}}), \text{GRD}_{v_{2}}(m_{v_{2}})$ \;
		$p_{v_{1}}^{G}, p_{v_{2}}^{G}~\leftarrow~g^{G}_{\theta}(\mathcal{R}^{{G}}_{v_{1}}), g^{G}_{\xi}(\mathcal{R}^{{G}}_{v_{2}})$ \;
		$\mathcal{L}_\text{GRCM} ~\leftarrow~ \frac{1}{N}\sum_{i=1}^{N}|| q^{G}_{\theta}(p_{v_{1}}^{G}) -  p_{v_{2}}^{G}|| \cdot M^{G}$ \;
		\KwRet $\mathcal{L}_\text{GRCM}$
	}
	
	\SetKwFunction{FMain}{Main}
	\Fn{\FMain}{
		\For{iter $\leq$ K} {
			$x, t_{1}, t_{2} \sim \mathbf{X}, \mathbf{T}, \mathbf{T}$
			$v_{1}, v_{2} ~\leftarrow~ t_{1}(x), t_{2}(x)$
			$m_{v_{1}},m_{v_{2}}  ~\leftarrow~ f_{\theta}(v_{1}),  f_{\xi}(v_{2})$
			$\mathcal{L}_\text{LRCM} ~\leftarrow~ \text{LRCM}(m_{v_{1}}, m_{v_{2}})$
			$\mathcal{L}_\text{GRCM} ~\leftarrow~ \text{GRCM}(m_{v_{1}}, m_{v_{2}})$
			$\mathcal{L}_\text{RCM} ~\leftarrow~ \mathcal{L}_\text{LRCM} + \mathcal{L}_\text{GRCM}$
			$\theta ~\leftarrow~ \text{Optimize}(\theta, \mathcal{L}_\text{RCM})$
			$\xi ~\leftarrow~ \tau\xi + (1 - \tau)\theta$
		}
		\KwRet $f_{\theta}$\;
	}
	\caption{Pseudo code of RCM.}
	\label{alg}
\end{algorithm}

\subsection{The parameters of DocReL}
The total amount parameters of DocReL are 142M. 
The output size of our ROI align is $1 \times 1$, so the input feature size of projection after ROI align is $128\times1\times1=128$. 
In comparison, The output feature size produced by~\cite{li2021structext} is $4 \times 64$, thus the input feature size of projection is $128\times4\times64=32768$.
Our implementation saves 24M parameters.

\subsection{Selection of the Basic Contrastive Learning Paradigm}
Our RCM task is built as a self-supervised, contrastive learning task.
The motivation is basically two folded.
Firstly, as large-scale free VRDs~\cite{wang2021layoutreader,li2020docbank,zhong2019publaynet} are available, acquiring their corresponding relational annotations for supervision is infeasible.
In addition, the definition of relation varies and may even conflict in a single downstream mission.
Based on the above observation, RCM should be unconstrained by a specific definition of relations and contrastive learning is a reasonable choice.

We select BYOL~\cite{grill2020bootstrap} as the basic contrastive learning paradigm, which avoids the usage of the instance's negative views.
Since negative views alleviate the collapsed problem~\cite{chen2020simple}, it is unfortunately inapplicable for RCM.
To explain, the basic instance in RCM is an entity in VRDs.
VRDs usually contain multiple entities, and an augmented view of VRD modifies the multi-modal feature of all entities.
Hence, their pairwise relation modification is unforeseeable.
What is worse, as the relation is an implicit property that relies on the deduction, it is non-trivial to verify the validity of the augmented view since relations are hard to be visualized.
Hence, exploiting positive views is a natural choice for DocReL.

\subsection{Feature Extraction of Entities in Broader Scenarios}
The relational representations provided by DocReL rely on precise entity-level feature extraction.
Now mainstream relational understanding tasks like table structure recognition~\cite{qasim2019rethinking,raja2020table,liu2021show,long2021parsing,liu2021neural}, key information extraction~\cite{hong2021bros,zhang2020trie,hwang2020spatial} and reading order detection~\cite{wang2021layoutreader} are born with entity-level.
To explain, the text-fields acquired from preconditioned OCR results or parsing results from electronic formats like PDF or Microsoft Word are directly entity-level, and no post-processing is needed.
However, an entity may be a single word within sentences or paragraphs, and directly exploiting the feature extraction system is inappropriate.
To solve this problem, a post-processing module is required to annotate text tokens and further locate entities.
This task is usually named token-level entity labeling module~\cite{li2021structext} which already yields satisfactory results.
Based on this observation, it is easy to generalize DocReL to broader scenarios.

{\small
\bibliographystyle{ieee_fullname}
\bibliography{Reference}
}
\end{document}